\documentclass[twocolumn,10pt]{asme2e}

\newcommand{\be}{\begin{equation}}
\newcommand{\ee}{\end{equation}}
\newcommand{\beq}{\begin{equation}}
\newcommand{\eeq}{\end{equation}}
\newcommand{\bed}{\begin{displaymath}}
\newcommand{\eed}{\end{displaymath}}
\newcommand{\beqa}{\begin{eqnarray}}
\newcommand{\eeqa}{\end{eqnarray}}
\newcommand{\beqann}{\begin{eqnarray*}}
\newcommand{\eeqann}{\end{eqnarray*}}
\newcommand{\bseq}{\begin{subequation}}
\newcommand{\eseq}{\end{subequation}}

\newcommand{\mat}[2]{\left[ \begin{array}{#1} #2 \end{array} \right] }

\newcommand{\ba}{\begin{array}}
\newcommand{\ea}{\end{array}}

\usepackage{longtable}
\usepackage[dvips]{graphicx}
\usepackage{psfrag, epsfig}
\usepackage{array}
\usepackage{amsmath,amssymb, amsfonts}
\usepackage{subfigure}
\usepackage{multirow}
\usepackage{color}

\confshortname{IDETC/CIE 2010}
\conffullname{the ASME 2010 International Design Engineering Technical Conferences \&\\
              Computers and Information in Engineering Conference}

\confdate{15-18}
\confmonth{August}
\confyear{2010}
\confcity{Montreal, Quebec}
\confcountry{Canada}

\papernum{DETC2010-28650}

\title{Comparison of Planar Parallel Manipulator Architectures based on a Multi-objective Design Optimization Approach}

\author{Damien CHABLAT, \, St\'{e}phane CARO, \, Raza UR-REHMAN, \, Philippe WENGER  
    \affiliation{
    Institut de Recherche en Communications
    et Cybern\'etique de Nantes\\
    UMR CNRS n$^\circ$ 6597\\
    1 rue de la No\"e, 44321
    Nantes, France\\
    Email: \{chablat,\,caro,\,ur-rehman,\,wenger\}@irccyn.ec-nantes.fr
    }
}

\begin{document}

\maketitle    

\begin{abstract}
{\it This paper deals with the comparison of planar parallel manipulator architectures based on a multi-objective design optimization approach. The manipulator architectures are compared with regard to their mass in motion and their regular workspace size, i.e., the objective functions. The optimization problem is subject to constraints on the manipulator dexterity and stiffness. For a given external wrench, the displacements of the moving platform have to be smaller than given values throughout the obtained maximum regular dexterous workspace. The contributions of the paper are highlighted with the study of 3-\underline{\textsf{P}}\textsf{R}\textsf{R}, 3-\textsf{R}\underline{\textsf{P}}\textsf{R} and 3-\underline{\textsf{R}}\textsf{R}\textsf{R} planar parallel manipulator architectures, which are compared by means of their Pareto frontiers obtained with a genetic algorithm.}
\end{abstract}

\section*{INTRODUCTION}

The design of parallel kinematics machines is a complex subject. The fundamental problem is that their performance heavily depends on their geometry \cite{Hay2004} and the mutual dependency of almost all the performance measures. This makes the problem computationally complex and yields the traditional solution approaches inefficient. As reported in~\cite{Merlet2006}, since the performance of a parallel manipulator depends on its dimensions, the latter depend on the manipulator application(s). Furthermore, numerous design aspects contribute to the Parallel Kinematics Machine~(PKM) performance and an efficient design will be one that takes into account all or most of these design aspects.  This is an iterative process and an efficient design requires a lot of computational efforts and capabilities for mapping design parameters into design criteria, and hence turning out with a multiobjective design optimization problem. Indeed, the optimal geometric parameters of a PKM can be determined by means of a the resolution of a multiobjective optimization problem. The solutions of such a problem are non-dominated solutions, also called Pareto-optimal solutions. Therefore, design optimization of parallel mechanisms is a key issue for their development.

Several researchers have focused on the optimization problem of parallel mechanisms the last few years. They have come up either with mono- or multi-objective design optimization problems. For instance, Lou et al.~\cite{Lou2005,Lou2008} presented a general approach for the optimal design of parallel manipulators to maximize the volume of an effective regular-shaped workspace while subject to constraints on their dexterity. Hay and Snyman~\cite{Hay2004} considered the optimal design of parallel manipulators to obtain a prescribed workspace, whereas Ottaviano and Ceccarelli~\cite{Ottaviano2000,Ottaviano2001} proposed a formulation for the optimum design of 3-Degree-Of-Freedom~(DOF) spatial parallel manipulators for given position and orientation workspaces. They based their study on the static analysis and the singularity loci of a manipulator in order to optimize the geometric design of the Tsai manipulator for a given free-singularity workspace. Hao and Merlet~\cite{Hao2005} discussed a multi-criterion optimal design methodology based on interval analysis to determine the possible geometric parameters satisfying two compulsory requirements of the workspace and accuracy. Similarly, Ceccarelli et al.~\cite{Ceccarelli2005} dealt with the multi-criterion optimum design of both parallel and serial manipulators with the focus on the workspace aspects, singularity and stiffness properties. Gosselin and Angeles~\cite{Gosselin1988,Gosselin1989} analyzed the design of a 3-DOF planar and a 3-DOF spherical parallel manipulators by maximizing their workspace volume while paying attention to their dexterity. Pham and Chen~\cite{Pham2004} suggested maximizing the workspace of a parallel flexible mechanism with the constraints on a global and uniformity measure of manipulability. Stamper et al.~\cite{Stamper1997} used the global conditioning index based on the integral of the inverse condition number of the kinematic Jacobian matrix over the workspace in order to optimize a spatial 3-DOF translational parallel manipulator. Stock and Miller~\cite{Stock2003} formulated a weighted sum multi-criterion optimization problem with manipulability and workspace as two objective functions. Menon et al.~\cite{Menon2009} used the maximization of the first natural frequency as an objective function for the geometrical optimization of the parallel mechanisms. Similarly, Li et al.~\cite{Li2009} proposed dynamics and elastodynamics optimization of a 2-DOF planar parallel robot to improve the dynamic accuracy of the mechanism. They proposed a dynamic index to identify the range of natural frequency with different configurations. Krefft~\cite{Krefft2005} also formulated a multi-criterion elastodynamic optimization problem for parallel mechanisms while considering workspace, velocity transmission, inertia, stiffness and the first natural frequency as optimization objectives. Chablat and Wenger~\cite{Chablat2003} proposed an analytical approach for the architectural optimization of a 3-DOF translational parallel mechanism, named Orthoglide~3-axis, based on prescribed kinetostatic performance to be satisfied in a given Cartesian workspace. 

Most of the foregoing research works aimed to improve the performance of a given manipulator and the comparison of various architectures for a given application or performance has not been considered. In this paper, the mechanisms performance are improved over a regular shaped workspace for given specifications. As a result, we propose a methodology to deal with the multiobjective design optimization of PKMs. The size of the regular shaped workspace and the mass in motion of the mechanism are the objective functions of the optimization problem. Its constraints are determined based on the mechanism accuracy, assembly and the conditioning number of its kinematic Jacobian matrix. The proposed approach is applied to the optimal design of Planar Parallel Manipulators~(PPMs) with the same mobility and set of design parameters. The non-dominated solutions, also called Pareto-optimal solutions, are obtained by means of a genetic algorithm for the three architectures and finally a comparison is made between them.
\section*{MANIPULATORS UNDER STUDY}
\begin{figure*}[!ht]
\centering
\subfigure[3-\underline{\textsf{P}}RR~PPM]{
  \psfrag{O}[c][c][0.85]{$O$}			 \psfrag{P}[c][c][0.85]{$P$}
  \psfrag{A1}[c][c][0.85]{$A_1$}   \psfrag{A2}[c][c][0.85]{$A_2$}		\psfrag{A3}[c][c][0.85]{$A_3$}
  \psfrag{B1}[tc][c][0.85]{$B_1$}  \psfrag{B2}[c][c][0.85]{$B_2$}		\psfrag{B3}[c][c][0.85]{$B_3$}
  \psfrag{C1}[c][c][0.85]{$C_1$}   \psfrag{C2}[c][c][0.85]{$C_2$}		\psfrag{C3}[c][c][0.85]{$C_3$}
  \psfrag{r1}[c][c][0.85]{$\rho_1$}	\psfrag{r2}[c][c][0.85]{$\rho_2$}	\psfrag{r3}[c][c][0.85]{$\rho_3$}
  \psfrag{R}[c][c][0.85]{$R$}	\psfrag{r}[c][c][0.85]{$r$}	\psfrag{pi}[c][c][0.85]{$\phi$}
\includegraphics[scale=0.37]{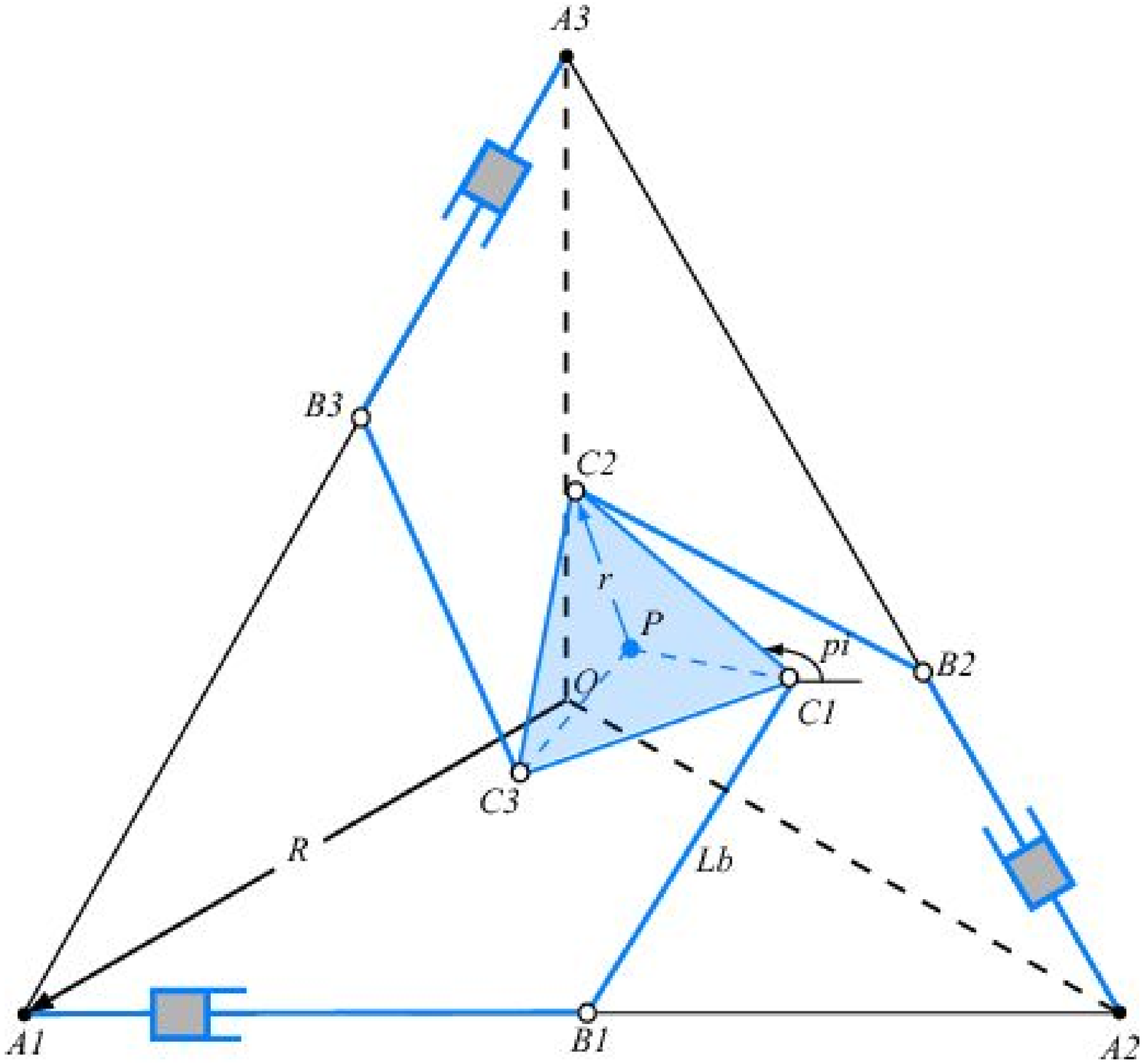}
\label{fig:PRR}
}
\subfigure[3-R\underline{\textsf{P}}R~PPM]{
  \psfrag{O}[c][c][0.85]{$O$}			 \psfrag{P}[c][c][0.85]{$P$}
  \psfrag{A1}[c][c][0.85]{$A_1$}   \psfrag{A2}[c][c][0.85]{$A_2$}		\psfrag{A3}[c][c][0.85]{$A_3$}
  \psfrag{B1}[tc][c][0.85]{$B_1$}  \psfrag{B2}[c][c][0.85]{$B_2$}		\psfrag{B3}[c][c][0.85]{$B_3$}
  \psfrag{C1}[c][c][0.85]{$C_1$}   \psfrag{C2}[c][c][0.85]{$C_2$}		\psfrag{C3}[c][c][0.85]{$C_3$}
  \psfrag{r1}[c][c][0.85]{$\rho_1$}	\psfrag{r2}[c][c][0.85]{$\rho_2$}	\psfrag{r3}[c][c][0.85]{$\rho_3$}
  \psfrag{R}[c][c][0.85]{$R$}	\psfrag{r}[c][c][0.85]{$r$}	\psfrag{pi}[c][c][0.85]{$\phi$}
	\includegraphics[scale=0.37]{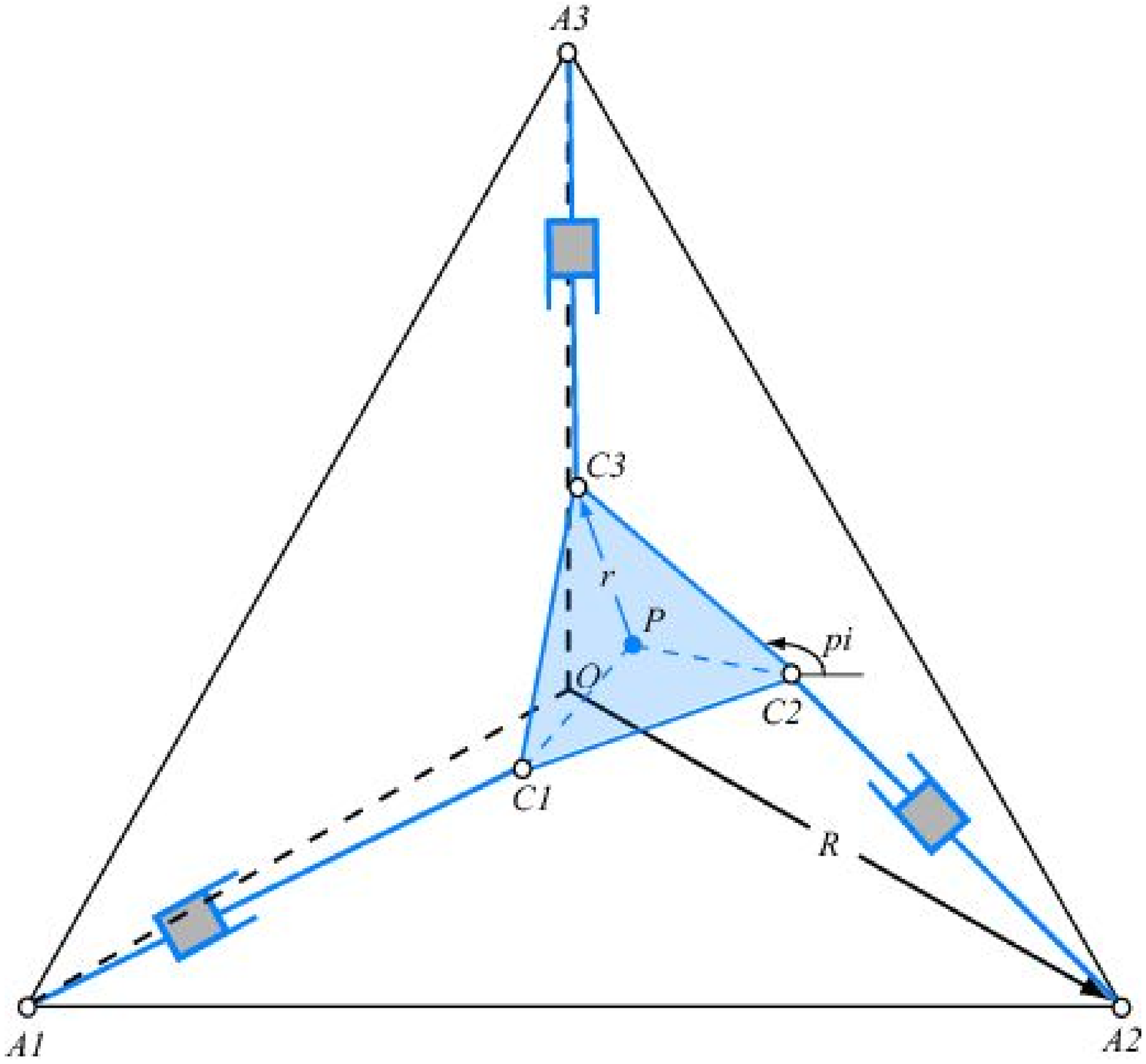}
\label{fig:RPR}
}
\subfigure[3-R\underline{\textsf{R}}R~PPM]{
  \psfrag{O}[c][c][0.85]{$O$}			 \psfrag{P}[c][c][0.85]{$P$}
  \psfrag{A1}[c][c][0.85]{$A_1$}   \psfrag{A2}[c][c][0.85]{$A_2$}		\psfrag{A3}[c][c][0.85]{$A_3$}
  \psfrag{B1}[tc][c][0.85]{$B_1$}  \psfrag{B2}[c][c][0.85]{$B_2$}		\psfrag{B3}[c][c][0.85]{$B_3$}
  \psfrag{C1}[c][c][0.85]{$C_1$}   \psfrag{C2}[c][c][0.85]{$C_2$}		\psfrag{C3}[c][c][0.85]{$C_3$}
  \psfrag{r1}[c][c][0.85]{$\rho_1$}	\psfrag{r2}[c][c][0.85]{$\rho_2$}	\psfrag{r3}[c][c][0.85]{$\rho_3$}
  \psfrag{R}[c][c][0.85]{$R$}	\psfrag{r}[c][c][0.85]{$r$}	\psfrag{pi}[c][c][0.85]{$\phi$} 
	\includegraphics[scale=0.37]{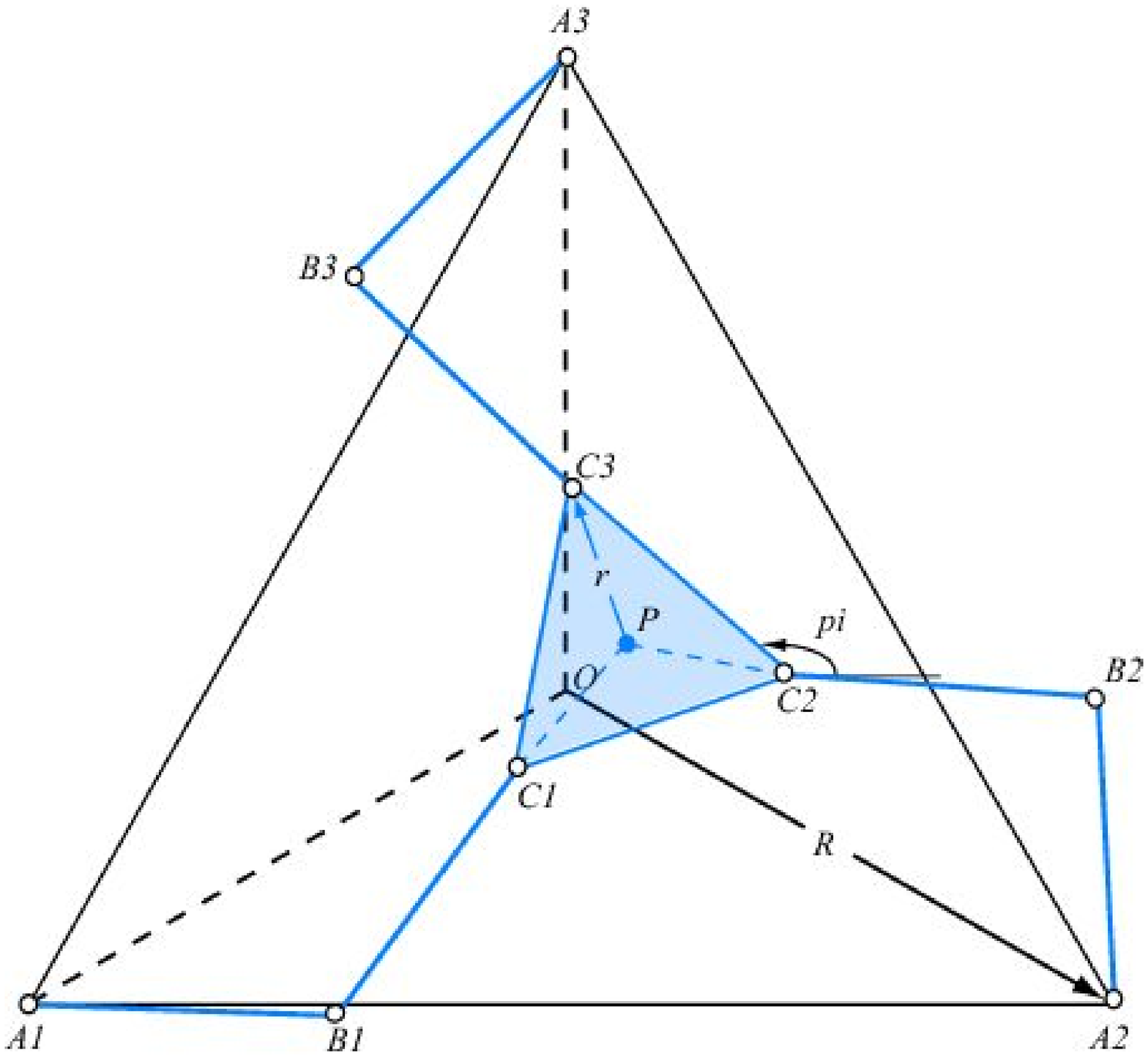}
\label{fig:RRR}
}
\caption{THE THREE PLANAR PARALLEL MANIPULATORS UNDER STUDY}
\end{figure*}

Figure~\ref{fig:PRR}--(c) illustrate the architectures of the planar parallel manipulators~(PPMs) under study, which are named 3-\underline{\textsf{P}}\textsf{R}\textsf{R}, 3-\textsf{R}\underline{\textsf{P}}\textsf{R} and 3-\underline{\textsf{R}}\textsf{R}\textsf{R}~PPMs, respectively. Other families of PPMs are described in~\cite{Merlet2006}. Here and throughout this paper, \textsf{R}, \textsf{P}, \underline{\textsf{R}} and \underline{\textsf{P}} denote revolute, prismatic, actuated revolute and actuated prismatic joints, respectively.  The manipulators under study are composed of a base and a moving platform~(MP) connected by means of three legs. Points $A_1$, $A_2$ and $A_3$, ($C_1$, $C_2$ and $C_3$, respectively) lie at the corners of a triangle, of which
point $O$ (point $P$, resp.) is the circumcenter. Each leg of the 3-\underline{\textsf{P}}\textsf{R}\textsf{R}~PPM is composed of a \textsf{P}, a \textsf{R} and a \textsf{R} joint in sequence. Each leg of the 3-\textsf{R}\underline{\textsf{P}}\textsf{R}~PPM is composed of a \textsf{R}, a \textsf{P} and a \textsf{R}~joint in sequence. Likewise, each leg of the 3-\underline{\textsf{R}}\textsf{R}\textsf{R}~PPM is composed of three \textsf{R} joints in sequence. The three~\textsf{P} joints of the 3-\underline{\textsf{P}}\textsf{R}\textsf{R} and the 3-\textsf{R}\underline{\textsf{P}}\textsf{R}~PPMs are actuated while the first \textsf{R}~joint of each leg of the 3-\underline{\textsf{R}}\textsf{R}\textsf{R}~PPM is actuated. 

$\mathcal{F}_b$ and $\mathcal{F}_p$ are the base and the moving platform frames of the manipulator. In the scope of this paper, $\mathcal{F}_b$ and $\mathcal{F}_p$ are supposed to be orthogonal. $\mathcal{F}_b$ is defined with the orthogonal dihedron $(\vec{Ox},\vec{Oy})$, point $O$ being its center and $\vec{Ox}$ parallel to segment $A_1A_2$. Likewise, $\mathcal{F}_p$ is defined with the orthogonal dihedron $(\vec{PX},\vec{PY})$, point $C$ being its center and $\vec{PX}$ parallel to segment $C_1C_2$. The manipulator MP pose, i.e., its position and its orientation, is
determined by means of the Cartesian coordinates vector ${\bf p}=\left[p_x,p_y\right]^T$
of operation point $P$ expressed in frame $\mathcal{F}_b$ and angle
$\phi$, namely, the angle between frames $\mathcal{F}_b$ and
$\mathcal{F}_p$.

The geometric parameters of the manipulators are defined as follows: (i)~$R$ is the circumradius of triangle $A_1A_2A_3$ of circumcenter $O$, i.e., $R=OA_i$; (ii)~$r$ is the circumradius of triangle $C_1C_2C_3$ of circumcenter $P$, i.e., $r=PC_i$, $i=1,\ldots,3$; (iii)~$L_b$ is the length of the intermediate links, i.e., $L_b=B_iC_i$ for the 3-\underline{\textsf{P}}\textsf{R}\textsf{R}~PPM. $L_b$ is also the maximum displacement of the prismatic joints of the 3-\textsf{R}\underline{\textsf{P}}\textsf{R}~PPM. Similarly, $L_b$ is the length of the two intermediate links of the 3-\underline{\textsf{R}}\textsf{R}\textsf{R}~PPM, i.e., $L_b=A_iB_i=B_iC_i$; (iv)~$r_j$ is the cross-section radius of the intermediate links; (v)~$r_p$: the cross-section radius of links of the moving platform, the latter being composed of three links.

\subsection*{Stiffness Modeling}
\label{Sec:SMatrix}
\begin{figure}[ht]
\centering
	\psfrag{BP}[t][t][0.7]{Base platform}%
	\psfrag{Rig}[t][t][0.7]{(Rigid)}%
	\psfrag{A}[t][t][0.7]{$A_c$}%
	\psfrag{R}[t][t][0.7]{$R$}%
	\psfrag{RL}[t][t][0.7]{Rigid body}%
	\psfrag{1DOF}[t][t][0.7]{1-dof}%
	\psfrag{6DOF}[t][t][0.7]{6-dof}%
	\psfrag{Sp}[t][t][0.7]{spring}%
	\psfrag{L}[t][t][0.7]{$L$}%
	\psfrag{r}[t][t][0.7]{$r$}%
	\psfrag{Ac}[t][t][0.7]{$A_c$}%
\includegraphics[scale=0.37]{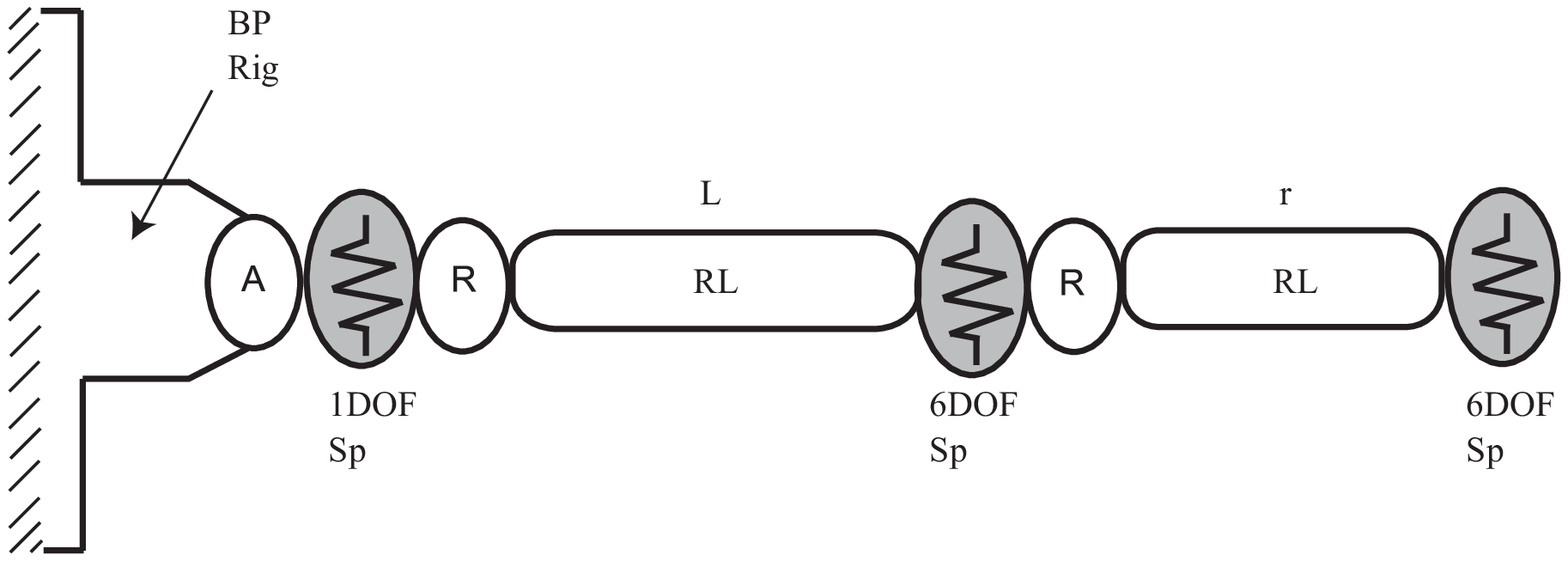}
\caption{FLEXIBLE MODEL OF THE 3-$\underline{P}RR$ PPM'S KINEMATIC CHAINS}
\label{Fig:FSM_3PRR}
\end{figure}

\begin{figure}[ht]
\centering
	\psfrag{BP}[t][t][0.7]{Base platform}%
	\psfrag{Rig}[t][t][0.7]{(Rigid)}%
	\psfrag{A}[t][t][0.7]{$A_c$}%
	\psfrag{R}[t][t][0.7]{$R$}%
	\psfrag{RL}[t][t][0.7]{Rigid body}%
	\psfrag{1DOF}[t][t][0.7]{1-dof}%
	\psfrag{6DOF}[t][t][0.7]{6-dof}%
	\psfrag{Sp}[t][t][0.7]{spring}%
	\psfrag{L}[t][t][0.7]{$L$}%
	\psfrag{r}[t][t][0.7]{$r$}%
	\psfrag{Ac}[t][t][0.7]{$A_c$}%
\includegraphics[scale=0.37]{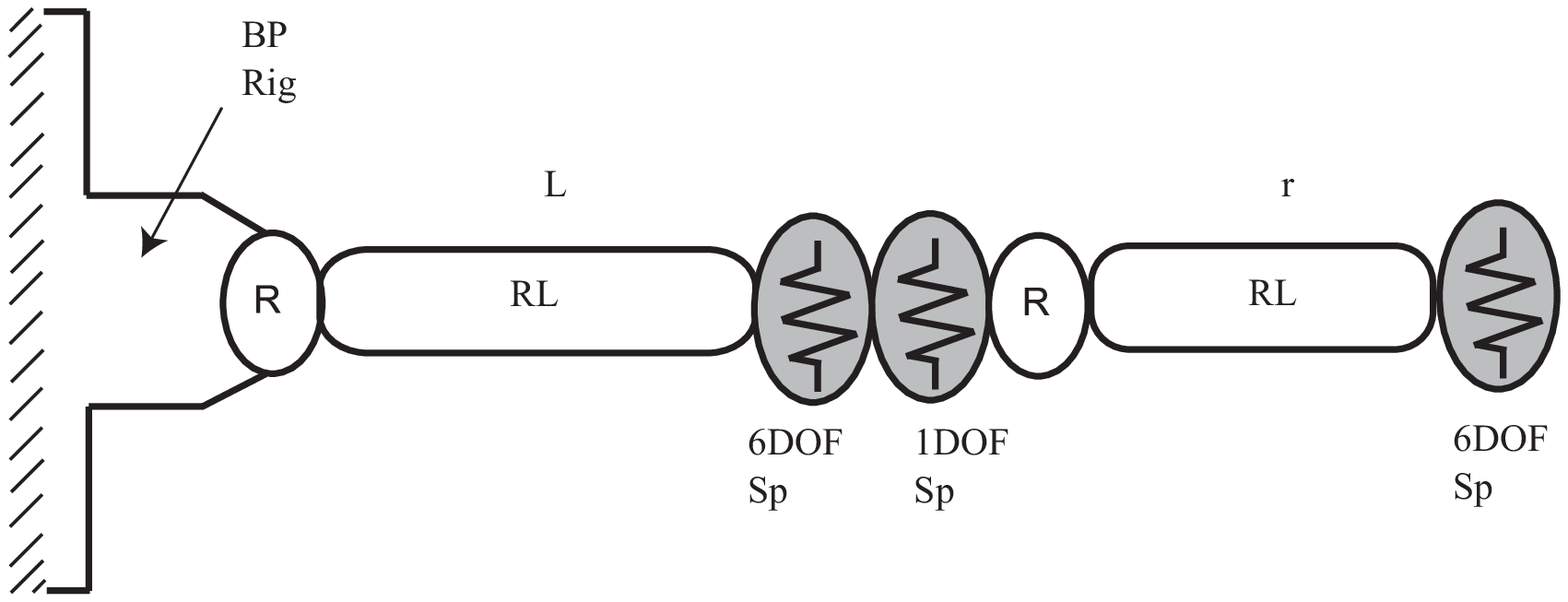}
\caption{FLEXIBLE MODEL OF THE 3-R$\underline{P}R$ PPM'S KINEMATIC CHAINS}
\label{Fig:FSM_3RPR}
\end{figure}

\begin{figure}[ht]
\centering
	\psfrag{BP}[t][t][0.7]{Base platform}%
	\psfrag{Rig}[t][t][0.7]{(Rigid)}%
	\psfrag{A}[t][t][0.7]{$A_c$}%
	\psfrag{R}[t][t][0.7]{$R$}%
	\psfrag{RL}[t][t][0.7]{Rigid body}%
	\psfrag{1DOF}[t][t][0.7]{1-dof}%
	\psfrag{6DOF}[t][t][0.7]{6-dof}%
	\psfrag{Sp}[t][t][0.7]{spring}%
	\psfrag{L}[t][t][0.7]{$L$}%
	\psfrag{r}[t][t][0.7]{$r$}%
	\psfrag{Ac}[t][t][0.7]{$A_c$}%
\includegraphics[scale=0.37]{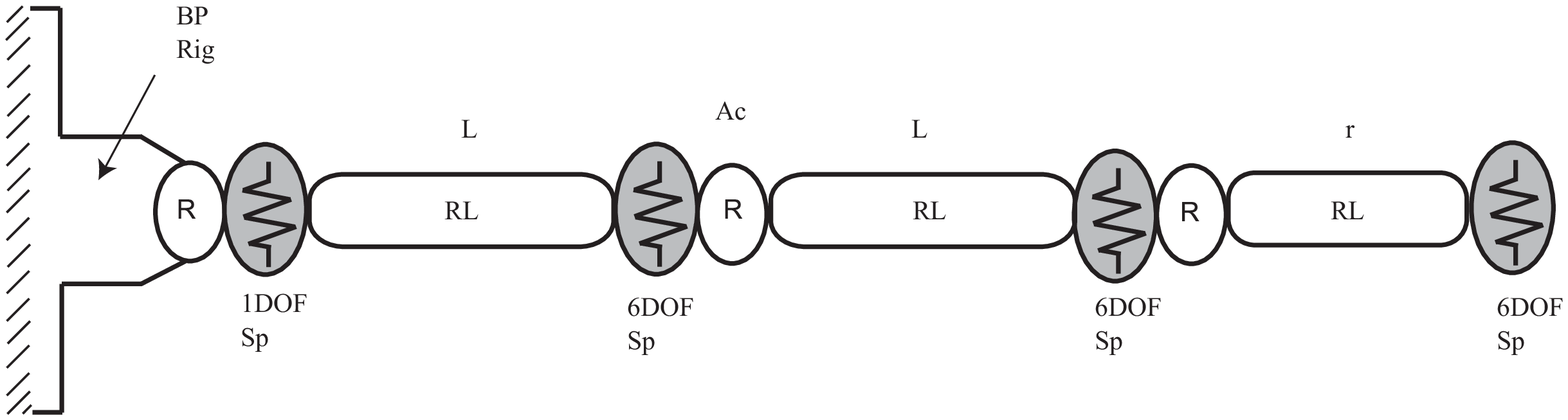}
\caption{FLEXIBLE MODEL OF THE 3-$\underline{R}RR$ PPM'S KINEMATIC CHAINS}
\label{Fig:FSM_3RRR}
\end{figure}

The stiffness models of the three manipulators under study are obtained by means of the refined lumped mass modeling described in~\cite{Pashkevich2009}. Figures~\ref{Fig:FSM_3PRR} to~\ref{Fig:FSM_3RRR} illustrate the flexible models of the legs of the 3-\underline{\textsf{P}}\textsf{R}\textsf{R}, 3-\textsf{R}\underline{\textsf{P}}\textsf{R} and 3-\underline{\textsf{R}}\textsf{R}\textsf{R}~PPMs, respectively. The actuator control loop compliance is described with a 1-dof virtual spring and the mechanical compliance of each link with a 6-dof virtual spring in each flexible model denoted $\theta_i$. Besides, the moving platform of the manipulators is supposed to be composed of three links of length $r$ connected to its geometric center~$P$. 

From Fig.~\ref{Fig:FSM_3PRR}, the flexible model of the legs of the 3-\underline{\textsf{P}}\textsf{R}\textsf{R}~PPM contains sequentially: (i)~a rigid link between the manipulator base and the $i^{th}$ actuated joint (part of the base platform) described by the constant homogeneous transformation matrix $\mathbf{T}^i_{Base}$; (ii)~a 1-dof actuated joint, defined by the homogeneous matrix function $\mathbf{V}_a(q^i_0)$ where $q^i_0$ is the actuated coordinate; (iii)~a 1-dof virtual spring describing the actuator mechanical stiffness, which is defined by the homogeneous matrix function $\mathbf{V}_{s1} \left( \theta^i_0 \right)$  where $\theta^i_0$  is the virtual spring coordinate corresponding to the translational spring; (iv)~a 1-dof passive $R$-joint at the beginning of the leg allowing one rotation angle $q^i_2$, which is described by the homogeneous matrix function $\mathbf{V}_{r1}(q^i_2)$; (v)~a rigid leg of length $L$ linking the foot and the movable platform, which is described by the constant homogeneous transformation matrix $\mathbf{T}^i_{L}$; (vi)~a 6-dof virtual spring describing the leg stiffness, which is defined by the homogeneous matrix function $\mathbf{V}_{s2} \left( \theta^i_1 \cdots \theta^i_6  \right)$, with $\theta^i_1,\; \theta^i_2,\; \theta^i_3$  and $\theta^i_4,\; \theta^i_5,\; \theta^i_6$  being the virtual spring coordinates corresponding to the spring translational and rotational deflections; (vii)~a 1-dof passive $R$-joint between the leg and the platform, allowing one rotation angle $q^i_3$, which is described by the homogeneous matrix function $\mathbf{V}_{r2}(q^i_3)$; (viii)~a rigid link of length $r$ from the manipulator leg to the geometric center of the mobile platform, which is described by the constant homogeneous transformation matrix $\mathbf{T}^i_{r}$; (ix)~a 6-dof virtual spring describing the stiffness of the moving platform, which is defined by the homogeneous matrix function $\mathbf{V}_{s3} \left( \theta^i_7 \cdots \theta^i_{12}  \right)$, $\theta^i_7,\; \theta^i_8,\; \theta^i_9$ and $\theta^i_{10},\; \theta^i_{11},\; \theta^i_{12}$ being the virtual spring coordinates corresponding to translational and rotational deflections of link $C_iP$; (x)~a homogeneous transformation matrix $\mathbf{T}^i_{End}$ that characterizes the rotation from the 6-dof spring associated with link $C_iP$ and the manipulator base frame.

As a result, the mathematical expression defining the end-effector location subject to variations in all above defined
coordinates of a single kinematic chain $i$ of the 3-\underline{\textsf{P}}\textsf{R}\textsf{R}~PPM takes the form:
\begin{eqnarray} \label{Eq:Ti2}
\mathbf{T}^i &=& \mathbf{T}^i_{Base} \mathbf{V}^i_a \left( q^i_0\right) \mathbf{V}_{s1} \left( \theta^i_0\right) \mathbf{V}_{r1}\left( q^i_1\right) \mathbf{T}^i_{L} \mathbf{V}_{s2} \left( \theta^i_1 \cdots \theta^i_6  \right) \nonumber \\ & & \mathbf{V}_{r2}(q^i_2) \mathbf{T}^i_{r} \mathbf{V}_{s3} \left( \theta^i_7 \cdots \theta^i_{12}  \right) \mathbf{T}^i_{End} 
\end{eqnarray}

Similarly, the mathematical expressions associated with the kinematic chains of the 3-\textsf{R}\underline{\textsf{P}}\textsf{R} and 3-\underline{\textsf{R}}\textsf{R}\textsf{R}~PPMs are obtained.

From \cite{Pashkevich2009}, the kinetostatic model of the i$th$ leg of the $X$-PPMs can be reduced to a system of two matrix equations, namely,
\begin{equation} \label{eq:kinetstatomodel}
	\mat{cc}{{\bf S}_{\theta|X}^i&{\bf J}_q^i\\{\bf J}_q^i&{\bf 0}_{2\times2}} \mat{c}{{\bf f}_i \\ \delta{\bf q}_i} = \mat{c}{\delta{\bf t}_i \\ {\bf 0}_2}
\end{equation}
where $X$ stands for 3-\underline{\textsf{P}}\textsf{R}\textsf{R}, 3-\textsf{R}\underline{\textsf{P}}\textsf{R} or 3-\underline{\textsf{R}}\textsf{R}\textsf{R}. The sub-matrix ${\bf S}_{\theta|X}^i={\bf J}_{\theta|X}^i {{\bf K}_{\theta|X}^{i}}^{-1} {{\bf J}_{\theta|X}^{i}}^T$ describes the spring compliance relative to the geometric center of the moving platform, and the sub-matrix ${\bf J}_q^i$ takes into account the passive joint influence on the moving platform motions. ${\bf J}_\theta^i$ is the Jacobian matrix related to the virtual springs and ${\bf J}_q^i$ is the one related to the passive joints. ${{\bf K}_{\theta|X}^{i}}^{-1}$ describes the compliance of the virtual springs.
\begin{subequations} 
\begin{eqnarray}
{\mathbf{K}^i_{\theta|3\underline{\textsf{P}}\textsf{R}\textsf{R}}}^{-1} &=& \mat{ccc}{
{\mathbf{K}^i_{act}}^{-1}	&	\mathbf{0}_{1\times6} & \mathbf{0}_{1\times6}\\
\mathbf{0}_{6\times1}	&	{\mathbf{K}^i_{link}}^{-1} &  \mathbf{0}_{6\times6} \\
\mathbf{0}_{6\times1}	&	\mathbf{0}_{6\times6} & {\mathbf{K}^i_{pf}}^{-1} } \label{Eq:KiPRR} \\
{\mathbf{K}^i_{\theta|3\textsf{R}\underline{\textsf{P}}\textsf{R}}}^{-1} &=& \mat{ccc}{
{\mathbf{K}^i_{link}}^{-1}	&	\mathbf{0}_{6\times1} & \mathbf{0}_{6\times6}\\
\mathbf{0}_{1\times6}	&	{\mathbf{K}^i_{act}}^{-1} &  \mathbf{0}_{1\times6} \\
\mathbf{0}_{6\times6}	&	\mathbf{0}_{6\times1} & {\mathbf{K}^i_{pf}}^{-1} } \label{Eq:KiRPR} \\
{\mathbf{K}^i_{\theta|3\underline{\textsf{R}}\textsf{R}\textsf{R}}}^{-1} &=& \mat{cccc}{
{\mathbf{K}^i_{act}}^{-1}	&	\mathbf{0}_{1\times6} & \mathbf{0}_{1\times6} & \mathbf{0}_{1\times6}\\
\mathbf{0}_{6\times1}	&	{\mathbf{K}^i_{link_1}}^{-1} & \mathbf{0}_{6\times6} & \mathbf{0}_{6\times6} \\
\mathbf{0}_{6\times1}	&	\mathbf{0}_{6\times6} & {\mathbf{K}^i_{link_2}}^{-1} & \mathbf{0}_{6\times6} \\
\mathbf{0}_{6\times1}	&	\mathbf{0}_{6\times6} & \mathbf{0}_{6\times6} & {\mathbf{K}^i_{pf}}^{-1}} \label{Eq:Ki}
\end{eqnarray}
\end{subequations} 
where $\mathbf{K}^i_{act}$ is the $1\times 1$ stiffness matrix of the i$th$ actuator, $\mathbf{K}^i_{link}$ is the $6\times 6$ stiffness matrix of the intermediate link for the 3-\underline{\textsf{P}}\textsf{R}\textsf{R} and 3-\textsf{R}\underline{\textsf{P}}\textsf{R}~PPMs while $\mathbf{K}^i_{link_1}$ and $\mathbf{K}^i_{link_2}$ are 
the $6\times 6$ stiffness matrices of the first and second intermediate links of the i$th$ leg of 3-\underline{\textsf{R}}\textsf{R}\textsf{R}~PPM. $\mathbf{K}^i_{pf}$ is the $6\times 6$ stiffness matrix of the i$th$ link of the moving platform. The compliance matrix of each link is expressed by means of the stiffness model of a cantilever beam, namely,
\begin{equation}
{\mathbf{K}^i_L}^{-1} = \mat{cccccc}{
\frac{L}{EA}& 		0 				& 			0 					& 			0 		& 			0					 & 			0 					\\
	0 		& \frac{L^3}{3EI_z}& 			0 					& 			0 		& 			0 				 & \frac{L^2}{2EI_z}\\
	0 		& 		0 						& \frac{L^3}{3EI_y} & 			0 		&-\frac{L^2}{2EI_y}& 		0 				\\
	0 		& 		0 						& 			0 					& \frac{L}{GI_x}& 		0					 & 			0 			\\
	0 		& 		0 						& -\frac{L^2}{2EI_y}& 			0 		&\frac{L}{EI_y} 	 & 			0 			\\
	0 		& \frac{L^2}{2EI_z} & 			0 					& 			0 		& 			0					 & \frac{L}{EI_z} 
  }
	\label{Eq:KiL}
\end{equation}
$L$ being the length of the corresponding link, $A$ is its the cross-sectional area, i.e., $A=\pi r_j^2$ for the links of the manipulators legs and $A=\pi r_p^2$ for the links of the moving platform. $I_y$ and $I_z$ are the polar moments of inertia about $y$ and $z$ axes, resp. $I_y=I_z=\pi r_j^4/4$ for the links of the manipulators legs and $I_y=I_z=\pi r_p^4/4$ for the links of the moving platform. $I_x\!=\!I_z+I_y$ is the polar moment of inertia about the longitudinal axis of the link. $E$ and $G$ are the Young and shear moduli of the material.

Accordingly, the Cartesian stiffness matrix ${\bf K}_i$ of the i$th$ leg defining the motion-to-force mapping is obtained from Eq.~(\ref{eq:kinetstatomodel}).
\begin{equation}
	{\bf f}_i = {\bf K}_i \, \delta{\bf t}_i
\end{equation}
with ${\bf f}_i$ being the wrench exerted on the i$th$ leg of the manipulator and at the geometric center of the moving platform while $\delta{\bf t}_i$ is the small-displacement screw of the moving-platform.

Finally, the Cartesian stiffness matrix ${\bf K}$ of the manipulator is found with a simple addition of the three ${\bf K}_i$ matrices, namely,
\begin{equation}
	{\bf K} = \sum_{i=1}^{3} {\bf K}_i
\end{equation}

\section*{MULTIOBJECTIVE OPTIMIZATION PROBLEM}
\label{Sec:MOOPF}
A multiobjective optimization problem~(MOOP) is formulated in this section in order to compare 3-\underline{\textsf{P}}\textsf{R}\textsf{R}, 3-\textsf{R}\underline{\textsf{P}}\textsf{R} and 3-\underline{\textsf{R}}\textsf{R}\textsf{R}~PPMs. In scope of this study, the manipulators are compared with regard to their mass in motion and their regular workspace size, i.e., the two objective functions of the MOOP, defined below. Moreover, the MOOP is subject to constraints on the manipulator dexterity and stiffness. It means that for a given external wrench, the displacements of the moving platform have to be smaller than given values throughout the obtained maximum regular dexterous workspace.
\subsection*{Objective Functions}
\subsubsection*{Mass in Motion of the Manipulators}
\noindent The components in motion of the manipulators are mainly their moving platform and the links of their legs. As a consequence, the mass in motion for the three PPMs under study is expressed as follows: 
\begin{subequations}
	\begin{eqnarray}
			m_{\underline{\textsf{P}}\textsf{R}\textsf{R}} &=& 3\,m_{link}+m_{pf} \label{eq:m3PRR} \\
			m_{\textsf{R}\underline{\textsf{P}}\textsf{R}} &=& 3\,m_{link}+m_{pf} \label{eq:m3RPR} \\
			m_{\underline{\textsf{R}}\textsf{R}\textsf{R}} &=& 6\,m_{link}+m_{pf} \label{Eq:m3RRR}
	\end{eqnarray}
\end{subequations} 
$m_{link}$ is the mass of links of the legs and are supposed to be the same while $m_{pf}$ is the mass of the moving platform. The mass of the prismatic or revolute actuators does not appear in Eqs.~(\ref{eq:m3PRR})-(c) as it is supposed to be fixed for the 3-\underline{\textsf{P}}\textsf{R}\textsf{R}~PPM and close to the base for the 3-\textsf{R}\underline{\textsf{P}}\textsf{R}~PPM.
\begin{subequations}
	\begin{eqnarray}
			m_{pf}=\pi\,r_p^2\,r\,\nu \label{eq:mpf} \\
			m_{link}=\pi\,r_j^2\,L\,\nu \label{eq:mlink}
	\end{eqnarray}
\end{subequations} 
where $\nu$ is the material density.

Finally, the first objective function of the MOOP is expressed as:
\begin{equation}
	f_1\left({\bf x}\right)=m_{X}\rightarrow \min
	\label{Eq:f1}
\end{equation}
${\bf x}$ being the vector of design variables, i.e., the geometric parameters of the manipulator at hand, and $X$ stands for 3-\underline{\textsf{P}}\textsf{R}\textsf{R}, 3-\textsf{R}\underline{\textsf{P}}\textsf{R} or 3-\underline{\textsf{R}}\textsf{R}\textsf{R}. 

\subsubsection*{Regular workspace size}
The quality of the manipulator workspace is of prime importance for the design of Parallel Kinematics Machines~(PKMs). It is partly characterized by its size and shape. Moreover, the lower the amount of 
singularities throughout the workspace, the better the workspace for continuous trajectory planning. The workspace optimization of parallel manipulators can usually be solved by means of two different formulations. The first formulation aims to design a manipulator whose workspace contains a prescribed workspace and the second one aims to design a manipulator, of which the workspace is as large as possible. However, maximizing the manipulator workspace may result in a poor design with regard to the manipulator dexterity and manipulability~\cite{Stamper1997,Wenger2000}. This problem can be solved by properly defining the constraints of the optimization problem. Here, the multiobjective optimization problem of PPMs is based on the formulation of workspace maximization, i.e, the determination of the optimum geometric parameters in order to maximize a regular-shaped workspace. 

\begin{figure}[ht]
\centering
  \psfrag{R}[c][c][0.8]{$R_w$}
  \psfrag{dp}[c][c][0.8]{$\Delta\phi$}
  \psfrag{C}[l][c][0.8]{$\left(x_c,\,y_c,\, \phi_c \right)$}
\includegraphics[scale=0.5]{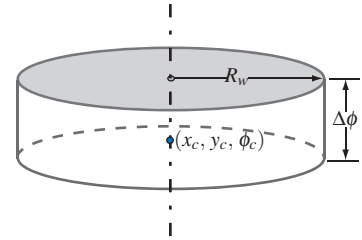}
\caption{A REGULAR-SHAPED WORKSPACE}
\label{Fig:Wspace}
\end{figure}

In the scope of the paper, the regular-shaped workspace is supposed to be a cylinder of radius $R_w$, for which at each point  a rotation range $\Delta\phi\!=\! 20 \,^{\circ}$ of the moving-platform about the $Z$-axis has to be reached. Figure~\ref{Fig:Wspace} illustrates such a regular-shaped workspace, whose $x_c$, $y_c$ and $\phi_c$ are its center coordinates and the rotation angle of the moving-platform of the manipulator in the home posture.

Consequently, in order to maximize the manipulator workspace, the second objective of the optimization problem can be written as:
\begin{equation}
	f_2\left(\mathbf{x}\right)=R_w\rightarrow \max
	\label{Eq:f2}
\end{equation}

\subsection*{Constraints of the Optimization Problem}
The constraints of the optimization problem deals with the geometric parameters, the dexterity and the accuracy of the manipulators. Moreover, the constraints have to be defined in order to obtain a singularity-free regular-shaped workspace.
\subsubsection*{Constraints on the Geometric Parameters}
For the three PPMs under study, the kinematic constraints are handled with their inverse kinematics. It means that the inverse kinematics is solved in order for the postures of the PPM to belong to the same working mode throughout the manipulator regular-shaped workspace. Besides, for the 3-\underline{\textsf{P}}\textsf{R}\textsf{R}~PPM, the lower and upper bounds of the prismatic lengths $\rho_i$ are defined such as $0 \leq \rho_i \leq \sqrt{3} R$ in order to avoid collisions. 
To obtain feasible displacements of the prismatic joints, the range of the 3-\textsf{R}\underline{\textsf{P}}\textsf{R}~PPM is defined such that $ L/2 \leq \rho_i \leq L$.
\subsubsection*{Constraint on the Manipulator Dexterity}
The manipulator dexterity is defined by the condition number of its kinematic Jacobian matrix. The {\em condition number} $\kappa_{F}({\bf M})$ of a $m \times n$
matrix ${\bf M}$, with $m \leq n$, based on the Frobenius norm is defined as follows
\begin{equation}
  \kappa_{F}({\bf M}) = \frac{1}{m} \sqrt{{\rm tr}({\bf M}^T{\bf M}){\rm tr}\left[({\bf M}^T{\bf M})^{-1}\right]}
\end{equation}
Here, the condition number is computed based on the Frobenius norm as the latter produces a condition number that is analytic in terms of the posture parameters whereas the 2-norm does not. Besides, it is much costlier to compute singular values than to compute matrix inverses.

The terms of the direct Jacobian matrix of the three PPMs under study are not
homogeneous as they do not have same units. Accordingly, its
condition number is meaningless. Indeed, its singular values cannot
be arranged in order as they are of different nature. However, from~\cite{Li:1990} and \cite{Paden:1988}, the Jacobian can be normalized
by means of a {\em normalizing length}. Later on,
the concept of {\em characteristic length} was introduced in~\cite{Ranjbaran:1995} in order to avoid the random choice of the normalizing
length. For instance, the previous concept was used in~\cite{Chablat2002} to analyze the kinetostatic performance of manipulators with multiple inverse kinematic solutions, and therefore to select their best {\em working mode}.

Accordingly, for the design optimization of the three PPMs, the minimum of the inverse condition number $\kappa^{-1} \left(\mathbf{J} \right)$ of the kinematic Jacobian matrix $\mathbf{J}$ is supposed to be higher than a prescribed value, say~0.1, throughout the regular-shaped workspace, for any rotation of its moving-platform, i.e.,
\begin{equation}
    min\left(\kappa^{-1} \left(\mathbf{J} \right) \right)\geq 0.1
	\label{Eq:C1}
\end{equation}
\subsubsection*{Constraints on the moving-platform pose errors}
\noindent The position and orientation errors on the moving-platform are evaluated by means of the stiffness models of the manipulators. Let $\left( \delta x,\;\delta y,\; \delta z \right)$ and $\left(\delta \phi_x,\; \delta \phi_y,\;\delta \phi_z\right)$ be the position and orientation errors of the moving-platform subject to external forces $\left(F_x,\; F_y,\; F_z \right)$ and torques $\left(\tau_z,\;\tau_y,\; \tau_z \right)$. 
The constraints on the pose errors on the moving-platform are defined as follows:
\begin{equation}
	\begin{array}{lll}
  \delta x  \leq  \delta x^{max} &  \delta y  \leq  \delta y^{max} \qquad  &  \delta z \leq \delta z^{max} \\
   \\
  \delta \phi_x \leq \delta \phi_x^{max} &   \delta \phi_y \leq \delta \phi_y^{max}  & \delta \phi_z \leq \delta \phi_z^{max}
    \end{array}
\label{Eq:C2PKM}
	\end{equation}
$\left(\delta x^{max},\;\delta y^{max},\;\delta z^{max} \right)$ being the maximum allowable position errors and $\left(\delta \phi_x^{max},\;\delta \phi_y^{max},\;\delta \phi_z^{max} \right)$ the maximum allowable orientation errors of the moving-platform. 
These accuracy constraints can be expressed in terms of the components of the mechanism stiffness matrix and the wrench applied to the moving-platform. Let us assume that the accuracy requirements are:
\begin{subequations}
	\begin{eqnarray}
    \sqrt{\delta x^2 + \delta y^2 } & \leq & 0.0001\, \textrm{m} \\
    \delta z & \leq & 0.001\, \textrm{m} \\
    \delta \phi_z  & \leq & 1\, \textrm{deg}
	\label{Eq:C2}
		\end{eqnarray}
\end{subequations}
If the moving-platform is subject to a wrench whose components are $\left\| F_{x,y}\right\|\!=\!F_z\!=\!100$\,N and $\tau_z\!=\!100$\,Nm, then the accuracy constraints can be expressed as:
\begin{subequations}
	\begin{eqnarray}
    k_{xy}^{min} & \geq &  \left\| F_{x,y}\right\|/ \sqrt{\delta x^2 + \delta y^2 } = 10^{6}~{\rm N.m\textsuperscript{-1}} \\
    k_{z}^{min} & \geq &  F_{z}/ \delta z = 10^{5}~{\rm N.m\textsuperscript{-1}} \\
    k_{\phi_z}^{min} & \geq & \tau_{z}/\delta \phi_z = \dfrac{10}{\pi/180}~{\rm N.m.rad\textsuperscript{-1}}
	\label{Eq:C3}
	\end{eqnarray}
\end{subequations}
\subsection*{Design Variables of the Optimization Problem}
\noindent Along with the above mentioned geometric parameters ($R,\, r,\, L_b$) of the PPMs, the radius $r_j$ of the circular-cross-section of the intermediate bars defined and the radius $r_p$ of the circular-cross-section of the platform bars are considered as design variables, also called decision variables. As a remainder, the moving-platform is supposed to composed of three circular bars of length $r$.

As there are three PPMs under study, the PPM type is another design variable that has to be taken into account. Let $d$ denote the PPM type: $d=1$ stands for the 3-\underline{\textsf{P}}\textsf{R}\textsf{R}~PPM; $d=2$ stands for the 3-\textsf{R}\underline{\textsf{P}}\textsf{R}~PPM; and $d=3$ stands for the 3-\underline{\textsf{R}}\textsf{R}\textsf{R}~PPM.

As a result, the optimization problem contains one discrete variable, i.e., $d$, and five continuous design variables, i.e., $R$, $r$, $L_b$, $r_j$ and $r_p$. Hence, the design variables vector ${\bf x}$ is given by:
\begin{equation}
		{\bf x}=\mat{cccccc}{d & R & r & L_b & r_j & r_p}^T
	\label{Eq:z}
\end{equation}
\subsection*{Formulation of the Optimization Problem}
The Multiobjective Design Optimization Problem of PPMs can be stated as: \emph{Find the optimum design variables ${\bf x}$ of PPMs in order to minimize the mass of the mechanism in motion  and to maximize its regular shaped workspace subject to geometric, kinematic and accuracy constraints.}

Mathematically, the problem can be written as:
\begin{align}
\label{Eq:MOO_PRR}  
{\rm minimize} &  \quad f_1(\mathbf{x})=m_{X}\\
{\rm maximize} & \quad 	f_2(\mathbf{x})=R_w \notag \\
& \notag \\
{\rm over} & \quad {\bf x}=\mat{cccccc}{d & R & r & L_b & r_j & r_p}^T \notag \\
& \notag \\
{\rm subject~to:} & \quad g_1: L_b + r \geq \dfrac{R}{2} \notag \\
& \quad g_2:0 <  \rho_i < \sqrt{3}R \notag \\
& \quad g_3: \kappa^{-1} \left(\mathbf{J} \right) \geq 0.1 \notag\\
& \quad g_4: k_{xy}^{min} \geq \dfrac{F_{x,y}}{\sqrt{\delta x^2 + \delta y^2 }} = 10^{6} \notag \\
& \quad g_5: k_{z}^{min} \geq \dfrac{F_{z}}{\delta z} = 10^{5} \notag \\
& \quad g_6: k_{\phi_z}^{min} \geq \dfrac{\tau_{z}}{\delta \phi_z} = \dfrac{10}{\pi/180}\notag\\
& \quad \mathbf{x}_{lb}\leq \mathbf{x} \leq \mathbf{x}_{ub} \notag 
\end{align}
where $\mathbf{x}_{lb}$ and $\mathbf{x}_{ub}$ are the lower and upper bounds of $\mathbf{x}$, respectively.

\section*{RESULTS AND DISCUSSIONS}
The multiobjective optimization problem~(\ref{Eq:MOO_PRR}) is solved by means of modeFRONTIER \cite{mFRONTIER} and by using its built-in multiobjective optimization algorithms. MATLAB code is incorporated in order to analyze the system and to get the numerical values for the objective functions and constraints that are analyzed in modeFRONTIER for their optimality and feasibility.
The lower and upper bounds of the design variables are given in Tab.~\ref{Tab:Xlb_ub}. The components of the PPMs are supposed to be made up of steel, of material density $d=7850$\,kg/m$^3$ and Young modulus $E=210\times10^9$\,N/m$^2$.
\begin{table}[!ht]		
		\caption{LOWER AND UPPER BOUNDS OF THE DESIGN VARIABLES}
	\centering
		\scalebox{.8}{  
		\begin{tabular}{lcccccc}
		\hline
		\hline
		Design Variable &  $d$ & $R$~[m] & $r$~[m] & $L_b$~[m] & $r_j$~[m] & $r_p$~[m]\\
		\hline
		Lower Bound & 1 & 0.5	&	0.5	&	0.5	&	0	&	0\\
		\hline
		Upper Bound & 3 & 4	& 4	&	4	&	0.1	&	0.1\\
		\hline
		\hline
		\end{tabular}
		}
	\label{Tab:Xlb_ub}
\end{table}
For each iteration, the regular-shaped workspace is evaluated for the corresponding design variables and a discretization of this workspace is performed. The constraints of the optimization problem are also evaluated at each grid point of the regular-shaped workspace to check whether they are satisfied or not.
\begin{table}[!ht]		
	\renewcommand{\arraystretch}{1.2}
	\caption{\emph{modeFRONTIER} ALGOTITHM PARAMETERS}
	\centering
			\scalebox{.8}{  
		\begin{tabular}{lc}
		\hline
		\hline
		Scheduler & MOGA-II\\
		\hline
		Number of iterations	&	 $200$ \\
		\hline
		Directional cross-over probability 	&	 $0.5$ \\
		\hline
		Selection probability&	 $0.05$ \\
		\hline
		Mutation probability	&	 $0.1$ \\
		\hline
		DNA~(DeoxyriboNucleic Acid) string & \multirow{2}{*}{$0.05$}  \\
    mutation ratio &   \\
    \hline
		DOE algorithm & Sobol\\
		\hline
		DOE number of designs &	 $30$ \\
		\hline
		Total number of iterations &	 $30\times 200 = 6000$ \\
		\hline
		\hline
		\end{tabular}
		}
	\label{Tab:DOE_PRR}
\end{table}
A multiobjective genetic algorithm~(MOGA) is used to solve MOOP~(\ref{Eq:MOO_PRR}) and to obtain the Pareto frontier in the plane defined by the mechanism mass and the workspace radius. \emph{modeFRONTIER} scheduler and Design Of Experiments~(DOE) parameters are given in Tab.~\ref{Tab:DOE_PRR}. MATLAB is used to evaluate each individual of the current population (generated by the \emph{modeFRONTIER} scheduler). \emph{MATLAB} returns the output variables that are analyzed by \emph{modeFRONTIER} for the feasible solutions according to the given constraints. At the end, the Pareto-optimal solutions are obtained from the generated feasible solutions.
\begin{figure}[ht]
\centering
%
%
%
\psfrag{s01}[t][t][1]{\color[rgb]{0,0,0}\setlength{\tabcolsep}{0pt}\begin{tabular}{c}Mass $m_t$ [kg]\end{tabular}}%
\psfrag{s02}[b][b][1]{\color[rgb]{0,0,0}\setlength{\tabcolsep}{0pt}\begin{tabular}{c}Workspace Radius $R_w$ [m]\end{tabular}}%
\psfrag{s03}[b][b][1]{\color[rgb]{0,0,0}\setlength{\tabcolsep}{0pt}\begin{tabular}{c} \end{tabular}}%
\psfrag{s05}[t][t][1]{\color[rgb]{0,0,0}\setlength{\tabcolsep}{0pt}\begin{tabular}{c}ID-I\end{tabular}}%
\psfrag{s06}[t][t][1]{\color[rgb]{0,0,0}\setlength{\tabcolsep}{0pt}\begin{tabular}{c}ID-II\end{tabular}}%
\psfrag{s07}[t][t][1]{\color[rgb]{0,0,0}\setlength{\tabcolsep}{0pt}\begin{tabular}{c}ID-III\end{tabular}}%

\psfrag{x01}[t][t][1.0]{0}%
\psfrag{x02}[t][t][1.0]{0.1}%
\psfrag{x03}[t][t][1.0]{0.2}%
\psfrag{x04}[t][t][1.0]{0.3}%
\psfrag{x05}[t][t][1.0]{0.4}%
\psfrag{x06}[t][t][1.0]{0.5}%
\psfrag{x07}[t][t][1.0]{0.6}%
\psfrag{x08}[t][t][1.0]{0.7}%
\psfrag{x09}[t][t][1.0]{0.9}%
\psfrag{x10}[t][t][1.0]{0.9}%
\psfrag{x11}[t][t][1.0]{1}%
\psfrag{x12}[t][t][1.0]{0}%
\psfrag{x13}[t][t][1.0]{200}%
\psfrag{x14}[t][t][1.0]{400}%
\psfrag{x15}[t][t][1.0]{600}%
\psfrag{x16}[t][t][1.0]{800}%
\psfrag{x17}[t][t][1.0]{1000}%
\psfrag{x18}[t][t][1.0]{1200}%
\psfrag{x19}[t][t][1.0]{1400}%
\psfrag{x20}[t][t][1.0]{1600}%
%
\psfrag{v01}[r][r][1.0]{0}%
\psfrag{v02}[r][r][1.0]{0.1}%
\psfrag{v03}[r][r][1.0]{0.2}%
\psfrag{v04}[r][r][1.0]{0.3}%
\psfrag{v05}[r][r][1.0]{0.4}%
\psfrag{v06}[r][r][1.0]{0.5}%
\psfrag{v07}[r][r][1.0]{0.6}%
\psfrag{v08}[r][r][1.0]{0.7}%
\psfrag{v09}[r][r][1.0]{0.8}%
\psfrag{v10}[r][r][1.0]{0.9}%
\psfrag{v11}[r][r][1.0]{1}%
\psfrag{v12}[r][r][1.0]{0}%
\psfrag{v13}[r][r][1.0]{0.2}%
\psfrag{v14}[r][r][1.0]{0.4}%
\psfrag{v15}[r][r][1.0]{0.6}%
\psfrag{v16}[r][r][1.0]{0.8}%
\psfrag{v17}[r][r][1.0]{1}%
\psfrag{v18}[r][r][1.0]{1.2}%
\psfrag{v19}[r][r][1.0]{1.4}%
\psfrag{v20}[r][r][1.0]{1.6}%
\psfrag{v21}[r][r][1.0]{1.8}%
%
%

\includegraphics[width=\linewidth]{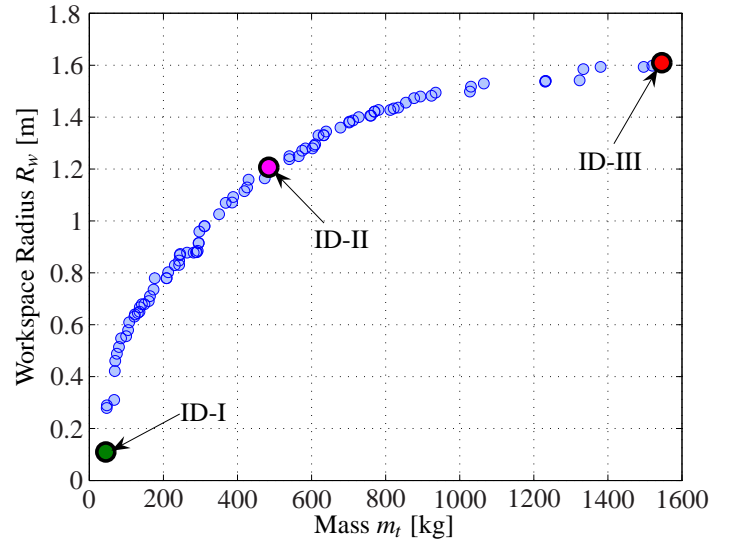}
\caption[Pareto frontier for 3-PRR design]{PARETO FRONTIER OF MOOP~(\ref{Eq:MOO_PRR})}
\label{Fig:Pareto_PRR}
\end{figure}	

\begin{figure*}[!ht]
\centering
\subfigure[ID--I]{
  \psfrag{WS}[l][c][0.7]{Regular Workspace}
	\includegraphics[scale=0.4]{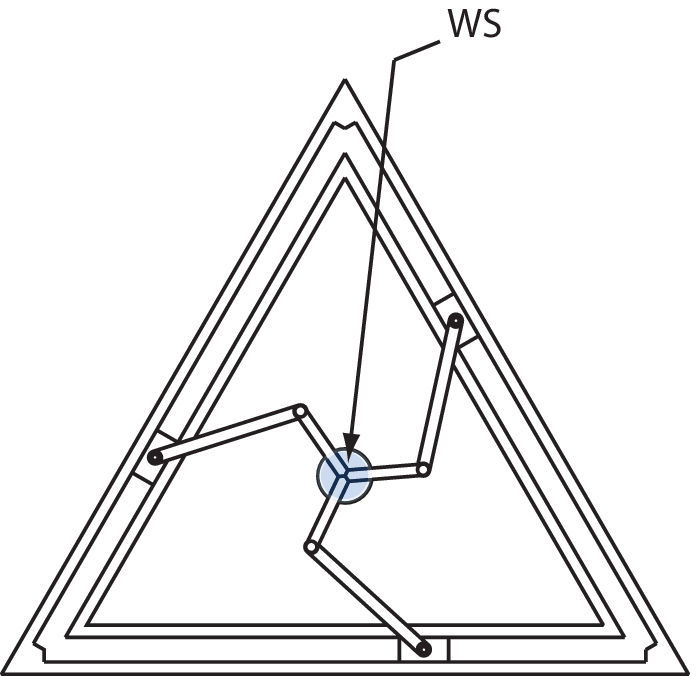}
\label{fig:ID1}
}
\subfigure[ID--II]{
  \psfrag{WS}[l][c][0.7]{Regular Workspace}
	\includegraphics[scale=0.4]{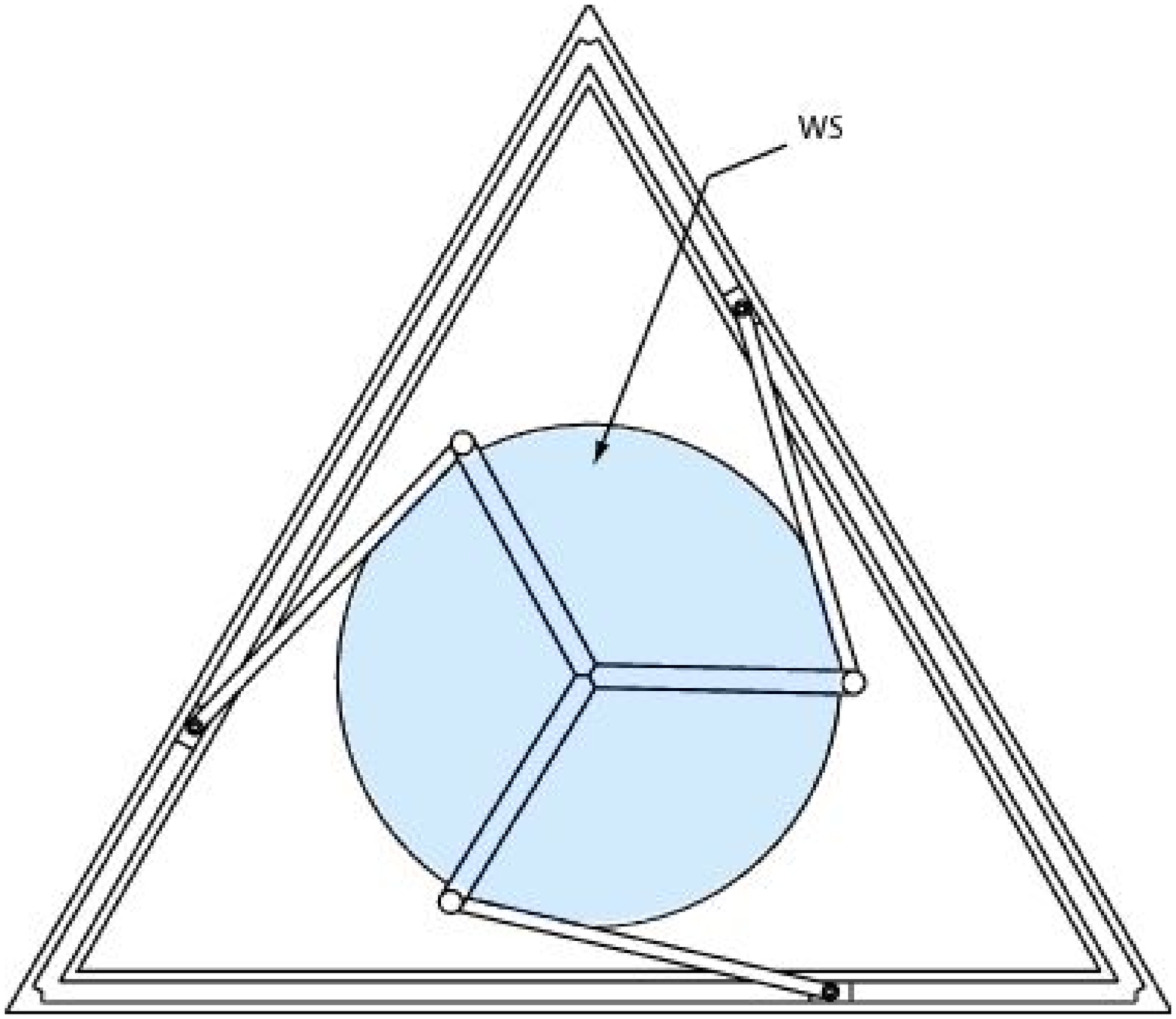}
\label{fig:ID3}
}
\subfigure[ID--III]{
  \psfrag{WS}[l][c][0.7]{Regular Workspace}
	\includegraphics[scale=0.4]{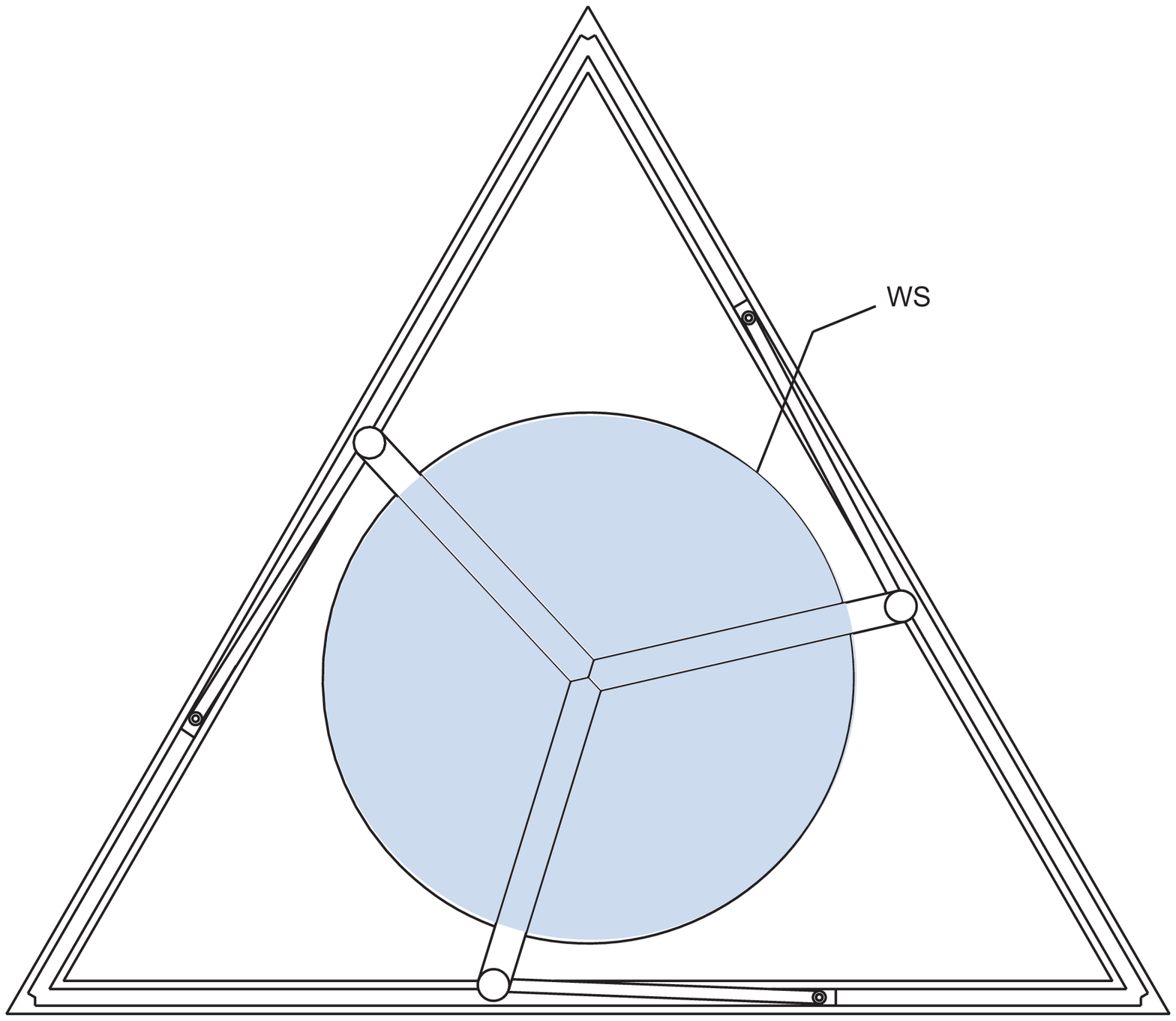}
\label{fig:ID2}
}
\caption{CAD DESIGNS OF THREE PARETO-OPTIMAL SOLUTIONS OF MOOP~(\ref{Eq:MOO_PRR})}
\label{Fig:ParetoDesigns}
\end{figure*}

The Pareto frontier, solution of MOOP~(\ref{Eq:MOO_PRR}), is depicted in Fig.~\ref{Fig:Pareto_PRR} whereas the design parameters and the corresponding objective functions for two extreme and one intermediate Pareto optimal solutions, as shown in Fig.~\ref{Fig:Pareto_PRR}, are given in Tab.~\ref{Tab:Pareto_PRR}. The CAD designs illustrating the three foregoing solutions are also shown in Fig.~\ref{Fig:ParetoDesigns}.

\begin{figure}[ht]
\centering
\input{Comparaison}
\includegraphics[width=\linewidth]{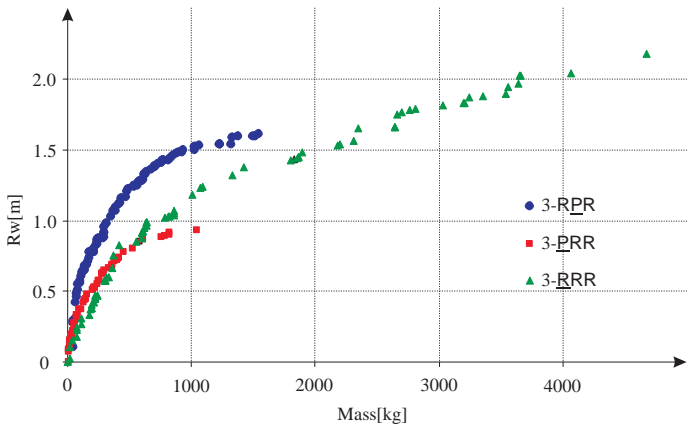}
\caption{PARETO FRONTIERS ASSOCIATED WITH THE 3-\underline{\textsf{P}}\textsf{R}\textsf{R}, 3-\textsf{R}\underline{\textsf{P}}\textsf{R}, AND 3-\underline{\textsf{R}}\textsf{R}\textsf{R}
PLANAR PARALLEL MANIPULATOR ARCHITECTURES}
\label{Fig:Comparaison}
\end{figure}

It appears that all Pareto-optimal solutions of MOOP~(\ref{Eq:MOO_PRR}) are 3-\underline{\textsf{P}}\textsf{R}\textsf{R}~PPMs. Accordingly, Fig.~\ref{Fig:Comparaison} illustrates the Pareto Frontiers associated with the three planar parallel manipulator architectures. It is noteworthy that the Pareto-optimal solutions associated with the 3-\underline{\textsf{P}}\textsf{R}\textsf{R}~PPM architectures are better than the Pareto-optimal solutions associated with the 3-\textsf{R}\underline{\textsf{P}}\textsf{R} and 3-\underline{\textsf{R}}\textsf{R}\textsf{R}~PPM architectures.

\begin{table*}[htbp]			
	\renewcommand{\arraystretch}{1.3}
	\centering
		\caption[Three Pareto optimal solutions]{THREE PARETO OPTIMAL SOLUTIONS}
	\scalebox{.8}{ 
		\begin{tabular}{c|cccccc|cc}
		\hline
		\hline
		\multirow{2}{12mm}{\centering Design ID} & \multicolumn{6}{c|}{ Design Variables} &		\multicolumn{2}{c}{\centering Objectives}\\	 
		\cline{2-9}
				& $d$ & $R$~[m] &   $r$~[m] 	& $L_b$~[m]	&	$r_j$~[m] &	$r_p$~[m] &  $m_t$~[kg] &  $R_w$~[m] \\
		\hline
		I		& 1 & 1.412	&	0.319	&	0.620	&	0.026	&	0.023	&	44.5	 &	0.110 \\
		\hline
		II	& 1 &	3.066	&	1.283	&	1.896	&	0.036	&	0.056	&	484.8	 &	1.207 \\
		\hline
		III	& 1 & 3.872	&	1.947 &	1.977 &	0.039	&	0.096	&	1545.6 & 1.609 \\
		\hline
		\hline
		\end{tabular}
		}
	\label{Tab:Pareto_PRR}
\end{table*}

\begin{figure*}[!ht]
\centering
\subfigure[3-\underline{\textsf{P}}\textsf{R}\textsf{R}~PPM]{
	\psfrag{0}[t][t][0.7]{$0$}%
	\psfrag{0.2}[t][t][0.7]{$0.2$}%
	\psfrag{0.4}[t][t][0.7]{$0.4$}%
	\psfrag{0.6}[t][t][0.7]{$0.6$}%
	\psfrag{0.8}[t][t][0.7]{$0.8$}%
	\psfrag{1}[t][t][0.7]{$1$}%
	\psfrag{1.2}[t][t][0.7]{$1.2$}%
	\psfrag{1.4}[t][t][0.7]{$1.4$}%
	\psfrag{1.6}[t][t][0.7]{$1.6$}%
	\psfrag{1.8}[t][t][0.7]{$1.8$}%
	\psfrag{0.000}[t][t][0.7]{$0$}%
	\psfrag{0.500}[t][t][0.7]{$0.5$}%
	\psfrag{1.000}[t][t][0.7]{$1.0$}%
	\psfrag{1.500}[t][t][0.7]{$1.5$}%
	\psfrag{2.000}[t][t][0.7]{$2.0$}%
	\psfrag{2.500}[t][t][0.7]{$2.5$}%
	\psfrag{3.000}[t][t][0.7]{$3.0$}%
	\psfrag{3.500}[t][t][0.7]{$3.5$}%
	\psfrag{4.000}[t][t][0.7]{$4.0$}%
	\psfrag{4.500}[t][t][0.7]{$4.5$}%
	\psfrag{R}[t][t][0.7]{$R$}%
	\psfrag{r}[t][t][0.7]{$r$}%
	\psfrag{L}[t][t][0.7]{$L$}%
	\psfrag{0.00}[t][t][0.7]{$0$}%
	\psfrag{0.10}[t][t][0.7]{$0.1$}%
	\psfrag{0.20}[t][t][0.7]{$0.2$}%
	\psfrag{0.30}[t][t][0.7]{$0.3$}%
	\psfrag{0.40}[t][t][0.7]{$0.4$}%
	\psfrag{0.50}[t][t][0.7]{$0.8$}%
	\psfrag{0.60}[t][t][0.7]{$0.6$}%
	\psfrag{0.70}[t][t][0.7]{$0.7$}%
	\psfrag{0.80}[t][t][0.7]{$0.8$}%
	\psfrag{0.90}[t][t][0.7]{$0.9$}%
	\psfrag{1.00}[t][t][0.7]{$1.0$}%
	\psfrag{Unit}[t][t][0.7]{[m]}%
	\psfrag{RDW}[c][c][0.7]{$R_w$~[m]}%
	\includegraphics[width=6cm,height=3.5cm]{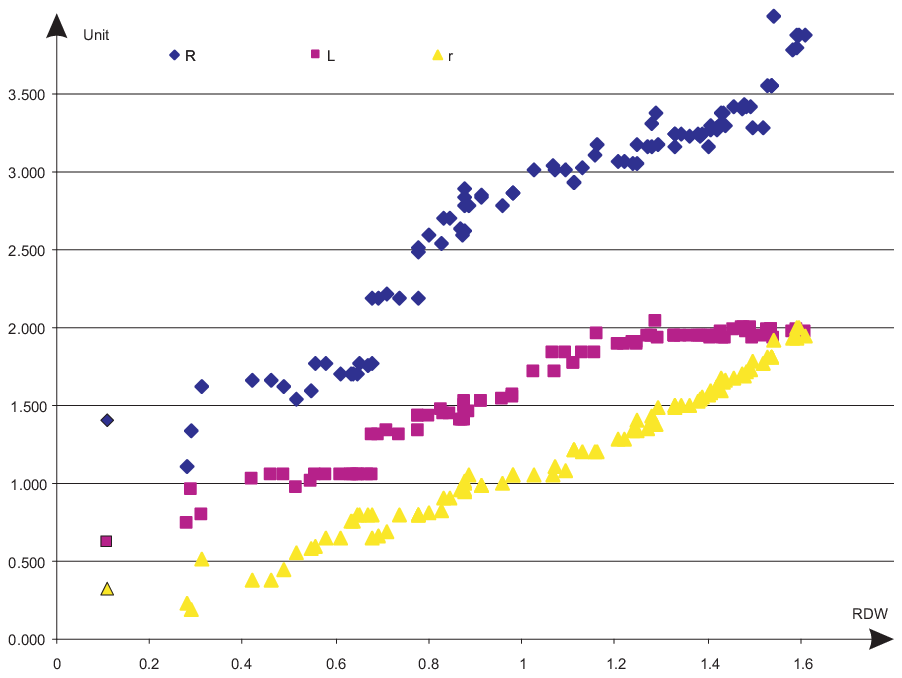}
\label{fig:DesVarRw3PRR1}
}
\subfigure[3-\textsf{R}\underline{\textsf{P}}\textsf{R}~PPM]{
	\psfrag{0}[t][t][0.7]{$0$}%
	\psfrag{0.2}[t][t][0.7]{$0.2$}%
	\psfrag{0.4}[t][t][0.7]{$0.4$}%
	\psfrag{0.6}[t][t][0.7]{$0.6$}%
	\psfrag{0.8}[t][t][0.7]{$0.8$}%
	\psfrag{1}[t][t][0.7]{$1$}%
	\psfrag{1.2}[t][t][0.7]{$1.2$}%
	\psfrag{1.4}[t][t][0.7]{$1.4$}%
	\psfrag{1.6}[t][t][0.7]{$1.6$}%
	\psfrag{1.8}[t][t][0.7]{$1.8$}%
	\psfrag{0.000}[t][t][0.7]{$0$}%
	\psfrag{0.500}[t][t][0.7]{$0.5$}%
	\psfrag{1.000}[t][t][0.7]{$1.0$}%
	\psfrag{1.500}[t][t][0.7]{$1.5$}%
	\psfrag{2.000}[t][t][0.7]{$2.0$}%
	\psfrag{2.500}[t][t][0.7]{$2.5$}%
	\psfrag{3.000}[t][t][0.7]{$3.0$}%
	\psfrag{3.500}[t][t][0.7]{$3.5$}%
	\psfrag{4.000}[t][t][0.7]{$4.0$}%
	\psfrag{4.500}[t][t][0.7]{$4.5$}%
	\psfrag{R}[t][t][0.7]{$R$}%
	\psfrag{r}[t][t][0.7]{$r$}%
	\psfrag{L}[t][t][0.7]{$L$}%
	\psfrag{0.00}[t][t][0.7]{$0$}%
	\psfrag{0.10}[t][t][0.7]{$0.1$}%
	\psfrag{0.20}[t][t][0.7]{$0.2$}%
	\psfrag{0.30}[t][t][0.7]{$0.3$}%
	\psfrag{0.40}[t][t][0.7]{$0.4$}%
	\psfrag{0.50}[t][t][0.7]{$0.8$}%
	\psfrag{0.60}[t][t][0.7]{$0.6$}%
	\psfrag{0.70}[t][t][0.7]{$0.7$}%
	\psfrag{0.80}[t][t][0.7]{$0.8$}%
	\psfrag{0.90}[t][t][0.7]{$0.9$}%
	\psfrag{1.00}[t][t][0.7]{$1.0$}%
	\psfrag{Unit}[t][t][0.7]{[m]}%
	\psfrag{RDW}[t][t][0.7]{$R_w$~[m]}%
	\includegraphics[width=6cm,height=3.5cm]{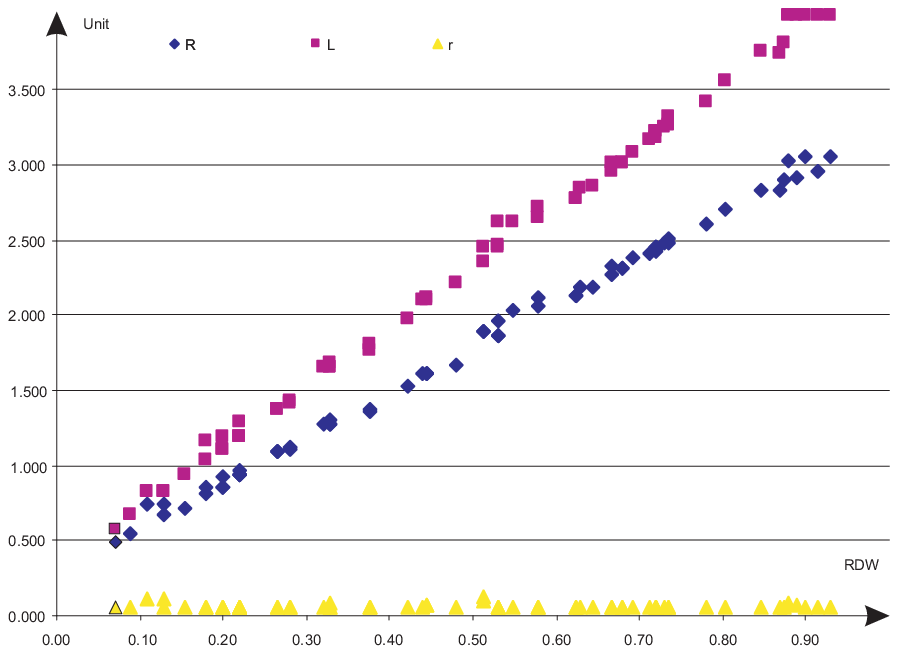}
\label{fig:DesVarRw3RPR1}
}
\subfigure[3-\underline{\textsf{R}}\textsf{R}\textsf{R}~PPM]{
	\psfrag{0}[t][t][0.7]{$0$}%
	\psfrag{0.2}[t][t][0.7]{$0.2$}%
	\psfrag{0.4}[t][t][0.7]{$0.4$}%
	\psfrag{0.6}[t][t][0.7]{$0.6$}%
	\psfrag{0.8}[t][t][0.7]{$0.8$}%
	\psfrag{1}[t][t][0.7]{$1$}%
	\psfrag{1.2}[t][t][0.7]{$1.2$}%
	\psfrag{1.4}[t][t][0.7]{$1.4$}%
	\psfrag{1.6}[t][t][0.7]{$1.6$}%
	\psfrag{1.8}[t][t][0.7]{$1.8$}%
	\psfrag{0.000}[t][t][0.7]{$0$}%
	\psfrag{0.500}[t][t][0.7]{$0.5$}%
	\psfrag{1.000}[t][t][0.7]{$1.0$}%
	\psfrag{1.500}[t][t][0.7]{$1.5$}%
	\psfrag{2.000}[t][t][0.7]{$2.0$}%
	\psfrag{2.500}[t][t][0.7]{$2.5$}%
	\psfrag{3.000}[t][t][0.7]{$3.0$}%
	\psfrag{3.500}[t][t][0.7]{$3.5$}%
	\psfrag{4.000}[t][t][0.7]{$4.0$}%
	\psfrag{4.500}[t][t][0.7]{$4.5$}%
	\psfrag{R}[t][t][0.7]{$R$}%
	\psfrag{r}[t][t][0.7]{$r$}%
	\psfrag{L}[t][t][0.7]{$L$}%
	\psfrag{0.00}[t][t][0.7]{$0$}%
	\psfrag{0.10}[t][t][0.7]{$0.1$}%
	\psfrag{0.20}[t][t][0.7]{$0.2$}%
	\psfrag{0.30}[t][t][0.7]{$0.3$}%
	\psfrag{0.40}[t][t][0.7]{$0.4$}%
	\psfrag{0.50}[t][t][0.7]{$0.8$}%
	\psfrag{0.60}[t][t][0.7]{$0.6$}%
	\psfrag{0.70}[t][t][0.7]{$0.7$}%
	\psfrag{0.80}[t][t][0.7]{$0.8$}%
	\psfrag{0.90}[t][t][0.7]{$0.9$}%
	\psfrag{1.00}[t][t][0.7]{$1.0$}%
	\psfrag{Unit}[t][t][0.7]{[m]}%
	\psfrag{RDW}[t][t][0.7]{$R_w$~[m]}%
	\includegraphics[width=6cm,height=3.5cm]{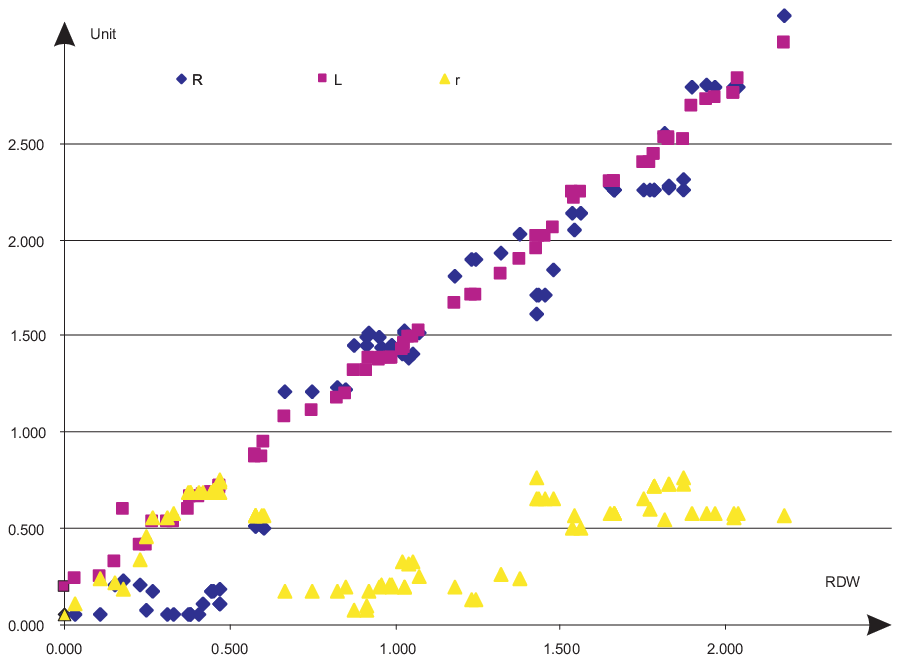}
\label{fig:DesVarRw3RRR1}
}
\caption{DESIGN VARIABLES $R$, $r$, $L_b$ AS A FUNCTION OF $R_w$ ALONG THE PARETO FRONTIER ASSOCIATED WITH THE MANIPULATOR AT HAND}
\label{fig:DesVarRwPPM1}
\end{figure*}

\begin{figure*}[!ht]
\centering
\subfigure[3-\underline{\textsf{P}}\textsf{R}\textsf{R}~PPM]{
	\psfrag{0}[t][t][0.7]{$0$}%
	\psfrag{0.2}[t][t][0.7]{$0.2$}%
	\psfrag{0.4}[t][t][0.7]{$0.4$}%
	\psfrag{0.6}[t][t][0.7]{$0.6$}%
	\psfrag{0.8}[t][t][0.7]{$0.8$}%
	\psfrag{0.10}[t][t][0.7]{$0.1$}%
	\psfrag{0.20}[t][t][0.7]{$0.2$}%
	\psfrag{0.30}[t][t][0.7]{$0.3$}%
	\psfrag{0.40}[t][t][0.7]{$0.4$}%
	\psfrag{0.50}[t][t][0.7]{$0.8$}%
	\psfrag{0.60}[t][t][0.7]{$0.6$}%
	\psfrag{0.70}[t][t][0.7]{$0.7$}%
	\psfrag{0.80}[t][t][0.7]{$0.8$}%
	\psfrag{0.90}[t][t][0.7]{$0.9$}%
	\psfrag{1.00}[t][t][0.7]{$1.0$}%
	\psfrag{1}[t][t][0.7]{$1$}%
	\psfrag{1.2}[t][t][0.7]{$1.2$}%
	\psfrag{1.4}[t][t][0.7]{$1.4$}%
	\psfrag{1.6}[t][t][0.7]{$1.6$}%
	\psfrag{1.8}[t][t][0.7]{$1.8$}%
	\psfrag{0.000}[t][t][0.7]{$0$}%
	\psfrag{0.020}[t][t][0.7]{$0.02$}%
	\psfrag{0.040}[t][t][0.7]{$0.04$}%
	\psfrag{0.060}[t][t][0.7]{$0.06$}%
	\psfrag{0.080}[t][t][0.7]{$0.08$}%
	\psfrag{0.100}[t][t][0.7]{$0.10$}%
	\psfrag{0.120}[t][t][0.7]{$0.12$}%
	\psfrag{0.000}[t][t][0.7]{$0$}%
	\psfrag{0.500}[t][t][0.7]{$0.5$}%
	\psfrag{1.000}[t][t][0.7]{$1.0$}%
	\psfrag{1.500}[t][t][0.7]{$1.5$}%
	\psfrag{2.000}[t][t][0.7]{$2.0$}%
	\psfrag{2.500}[t][t][0.7]{$2.5$}%
	\psfrag{0.01}[t][t][0.7]{$0.01$}%
	\psfrag{0.02}[t][t][0.7]{$0.02$}%
	\psfrag{0.03}[t][t][0.7]{$0.03$}%
	\psfrag{0.04}[t][t][0.7]{$0.04$}%
	\psfrag{0.05}[t][t][0.7]{$0.05$}%
	\psfrag{0.06}[t][t][0.7]{$0.06$}%
	\psfrag{0.07}[t][t][0.7]{$0.07$}%
	\psfrag{r_j}[t][t][0.7]{$r_j$}%
	\psfrag{r_p}[t][t][0.7]{$r_p$}%
	\psfrag{Unit}[t][t][0.7]{[m]}%
	\psfrag{RDW}[t][t][0.7]{$R_w$~[m]}%
	\includegraphics[width=6cm,height=3.5cm]{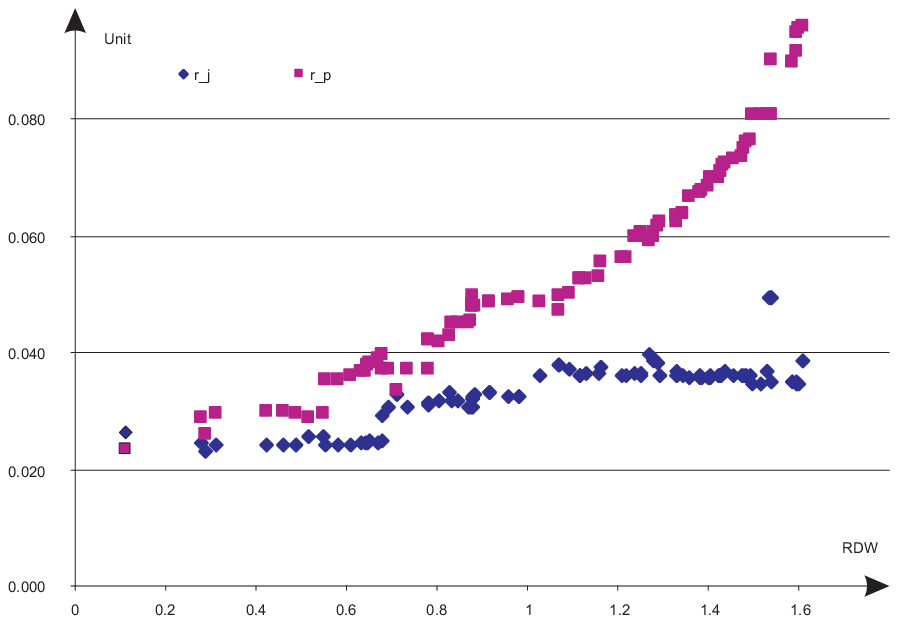}
\label{fig:DesVarRw3PRR2}
}
\subfigure[3-\textsf{R}\underline{\textsf{P}}\textsf{R}~PPM]{
	\psfrag{0}[t][t][0.7]{$0$}%
	\psfrag{0.2}[t][t][0.7]{$0.2$}%
	\psfrag{0.4}[t][t][0.7]{$0.4$}%
	\psfrag{0.6}[t][t][0.7]{$0.6$}%
	\psfrag{0.8}[t][t][0.7]{$0.8$}%
	\psfrag{0.10}[t][t][0.7]{$0.1$}%
	\psfrag{0.20}[t][t][0.7]{$0.2$}%
	\psfrag{0.30}[t][t][0.7]{$0.3$}%
	\psfrag{0.40}[t][t][0.7]{$0.4$}%
	\psfrag{0.50}[t][t][0.7]{$0.8$}%
	\psfrag{0.60}[t][t][0.7]{$0.6$}%
	\psfrag{0.70}[t][t][0.7]{$0.7$}%
	\psfrag{0.80}[t][t][0.7]{$0.8$}%
	\psfrag{0.90}[t][t][0.7]{$0.9$}%
	\psfrag{1.00}[t][t][0.7]{$1.0$}%
	\psfrag{1}[t][t][0.7]{$1$}%
	\psfrag{1.2}[t][t][0.7]{$1.2$}%
	\psfrag{1.4}[t][t][0.7]{$1.4$}%
	\psfrag{1.6}[t][t][0.7]{$1.6$}%
	\psfrag{1.8}[t][t][0.7]{$1.8$}%
	\psfrag{0.000}[t][t][0.7]{$0$}%
	\psfrag{0.020}[t][t][0.7]{$0.02$}%
	\psfrag{0.040}[t][t][0.7]{$0.04$}%
	\psfrag{0.060}[t][t][0.7]{$0.06$}%
	\psfrag{0.080}[t][t][0.7]{$0.08$}%
	\psfrag{0.100}[t][t][0.7]{$0.10$}%
	\psfrag{0.120}[t][t][0.7]{$0.12$}%
	\psfrag{0.000}[t][t][0.7]{$0$}%
	\psfrag{0.500}[t][t][0.7]{$0.5$}%
	\psfrag{1.000}[t][t][0.7]{$1.0$}%
	\psfrag{1.500}[t][t][0.7]{$1.5$}%
	\psfrag{2.000}[t][t][0.7]{$2.0$}%
	\psfrag{2.500}[t][t][0.7]{$2.5$}%
	\psfrag{0.01}[t][t][0.7]{$0.01$}%
	\psfrag{0.02}[t][t][0.7]{$0.02$}%
	\psfrag{0.03}[t][t][0.7]{$0.03$}%
	\psfrag{0.04}[t][t][0.7]{$0.04$}%
	\psfrag{0.05}[t][t][0.7]{$0.05$}%
	\psfrag{0.06}[t][t][0.7]{$0.06$}%
	\psfrag{0.07}[t][t][0.7]{$0.07$}%
	\psfrag{r_j}[t][t][0.7]{$r_j$}%
	\psfrag{r_p}[t][t][0.7]{$r_p$}%
	\psfrag{Unit}[t][t][0.7]{[m]}%
	\psfrag{RDW}[t][t][0.7]{$R_w$~[m]}%
		\includegraphics[width=6cm,height=3.5cm]{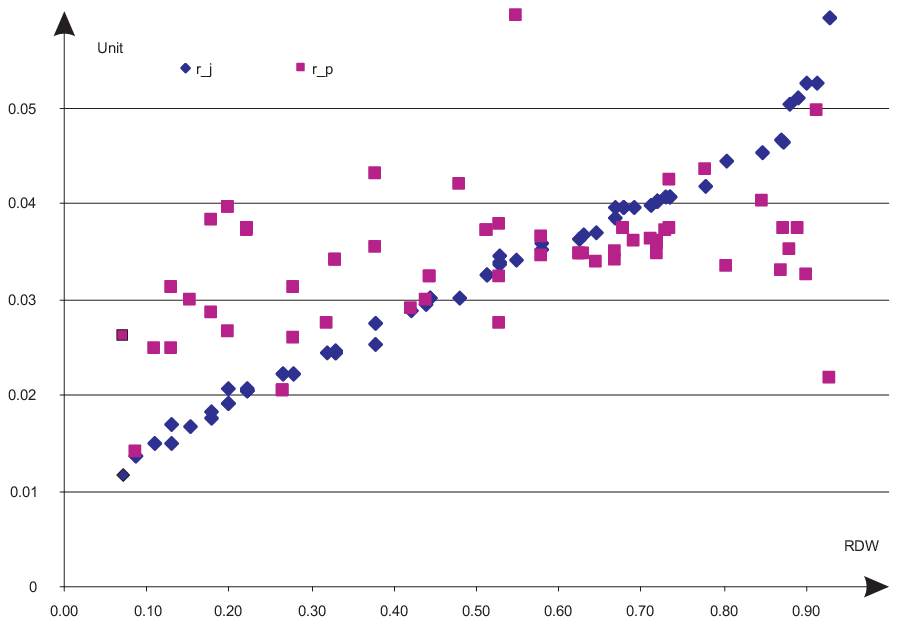}
\label{fig:DesVarRw3RPR2}
}
\subfigure[3-\underline{\textsf{R}}\textsf{R}\textsf{R}~PPM]{
	\psfrag{0}[t][t][0.7]{$0$}%
	\psfrag{0.2}[t][t][0.7]{$0.2$}%
	\psfrag{0.4}[t][t][0.7]{$0.4$}%
	\psfrag{0.6}[t][t][0.7]{$0.6$}%
	\psfrag{0.8}[t][t][0.7]{$0.8$}%
	\psfrag{0.10}[t][t][0.7]{$0.1$}%
	\psfrag{0.20}[t][t][0.7]{$0.2$}%
	\psfrag{0.30}[t][t][0.7]{$0.3$}%
	\psfrag{0.40}[t][t][0.7]{$0.4$}%
	\psfrag{0.50}[t][t][0.7]{$0.8$}%
	\psfrag{0.60}[t][t][0.7]{$0.6$}%
	\psfrag{0.70}[t][t][0.7]{$0.7$}%
	\psfrag{0.80}[t][t][0.7]{$0.8$}%
	\psfrag{0.90}[t][t][0.7]{$0.9$}%
	\psfrag{1.00}[t][t][0.7]{$1.0$}%
	\psfrag{1}[t][t][0.7]{$1$}%
	\psfrag{1.2}[t][t][0.7]{$1.2$}%
	\psfrag{1.4}[t][t][0.7]{$1.4$}%
	\psfrag{1.6}[t][t][0.7]{$1.6$}%
	\psfrag{1.8}[t][t][0.7]{$1.8$}%
	\psfrag{0.000}[t][t][0.7]{$0$}%
	\psfrag{0.020}[t][t][0.7]{$0.02$}%
	\psfrag{0.040}[t][t][0.7]{$0.04$}%
	\psfrag{0.060}[t][t][0.7]{$0.06$}%
	\psfrag{0.080}[t][t][0.7]{$0.08$}%
	\psfrag{0.100}[t][t][0.7]{$0.10$}%
	\psfrag{0.120}[t][t][0.7]{$0.12$}%
	\psfrag{0.000}[t][t][0.7]{$0$}%
	\psfrag{0.500}[t][t][0.7]{$0.5$}%
	\psfrag{1.000}[t][t][0.7]{$1.0$}%
	\psfrag{1.500}[t][t][0.7]{$1.5$}%
	\psfrag{2.000}[t][t][0.7]{$2.0$}%
	\psfrag{2.500}[t][t][0.7]{$2.5$}%
	\psfrag{0.01}[t][t][0.7]{$0.01$}%
	\psfrag{0.02}[t][t][0.7]{$0.02$}%
	\psfrag{0.03}[t][t][0.7]{$0.03$}%
	\psfrag{0.04}[t][t][0.7]{$0.04$}%
	\psfrag{0.05}[t][t][0.7]{$0.05$}%
	\psfrag{0.06}[t][t][0.7]{$0.06$}%
	\psfrag{0.07}[t][t][0.7]{$0.07$}%
	\psfrag{r_j}[t][t][0.7]{$r_j$}%
	\psfrag{r_p}[t][t][0.7]{$r_p$}%
	\psfrag{Unit}[t][t][0.7]{[m]}%
	\psfrag{RDW}[t][t][0.7]{$R_w$~[m]}%
	\includegraphics[width=6cm,height=3.5cm]{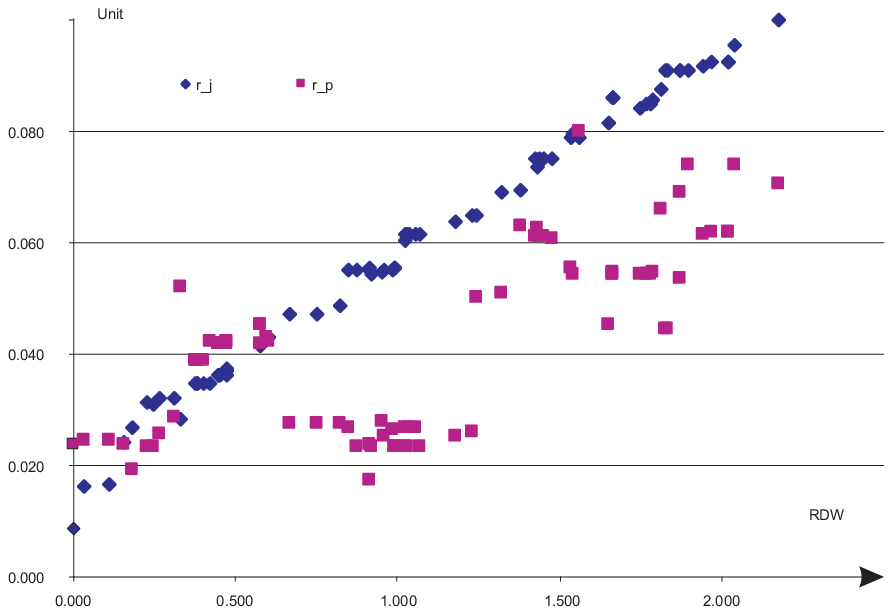}
\label{fig:DesVarRw3RRR2}
}
\caption{DESIGN VARIABLES $r_j$, $r_p$ AS A FUNCTION OF $R_w$ ALONG THE PARETO FRONTIER ASSOCIATED WITH THE MANIPULATOR AT HAND}
\label{fig:DesVarRwPPM2}
\end{figure*}

Figures~\ref{fig:DesVarRw3PRR1}--(c) and \ref{fig:DesVarRw3PRR2}--(c) show the evolution of the design variables as a function of $R_w$ along the Pareto Frontier associated with each PPM architecture. It is noteworthy that the higher $R_w$, the higher the design variables. It is apparent that the variations in variables $R$, $r$, $L_b$ and $r_j$ with respect to~(w.r.t.) $R_w$ are almost linear whereas the variations in $r_p$ w.r.t. $R_w$ is rather quadratic. This is due to the fact that the higher the size of the mechanism the higher the bending of the moving platform links whereas the intermediate links are mainly subjected to tension and compression.

\section*{CONCLUSIONS}
In this paper, the problem of dimensional synthesis of parallel kinematics machines was addressed. A multiobjective design optimization problem was formulated in order to determine optimum structural and geometric parameters of any parallel kinematics machine. The proposed approach is similar to that used in \cite{Altuzarra2009} but we took into account the mass and the regular workspace instead of considering the entire volume of the manipulator. The proposed approach was applied to the optimum design of three planar parallel manipulators with the aim to minimize the mass in motion of the mechanism and to maximize its regular shaped workspace. Other performance indices can be used as constraints. However, they cannot necessarily be used as objective functions as the latter are usually formulated as a sum of an index over all the manipulator workspace. As another constraint, we could use the collisions between the legs of the manipulator.

\bibliographystyle{unsrt}
\bibliography{JabRefRazaThesis}

\end{document}